%% file: article.tex
\documentclass[pdflatex,sn-nature]{sn-jnl}

\input{preamble}

\theoremstyle{thmstyleone}%
\theoremstyle{thmstyletwo}%

\theoremstyle{thmstylethree}%

\begin{document}

\title[Article Title]{Generative Flow Networks for Model Adaptation in Digital Twins of Natural Systems}


\author*[1]{\fnm{Pascal} \sur{Archambault}}\email{pascal.archambault@umontreal.ca}

\author[1]{\fnm{Houari} \sur{Sahraoui}}\email{sahraouh@iro.umontreal.ca}

\author[1]{\fnm{Eugene} \sur{Syriani}}\email{syriani@iro.umontreal.ca}

\affil*[1]{\orgdiv{\textit{DIRO}}, \orgname{Université de Montréal}, \state{Quebec}, \country{Canada}}

\abstract{Digital twins of natural systems must remain aligned with physical systems that evolve over time, are only partially observed, and are typically modeled by mechanistic simulators whose parameters cannot be measured directly. In such settings, model adaptation is naturally posed as a simulation-based inference problem. However, sparse and indirect observations often fail to identify a unique and optimal calibration, leaving several simulator parameterizations compatible with the available evidence. This article presents a GFlowNet-based approach to model adaptation for digital twins of natural systems. We formulate adaptation as a generative modeling problem over complete simulator configurations, so that plausible parameterizations can be sampled with probability proportional to a reward derived from agreement between simulated and observed behavior. Using a controlled environment agriculture case study based on a mechanistic tomato model, we show that the learned policy recovers dominant regions of the adaptation landscape, retrieves strong calibration hypotheses, and preserves multiple plausible configurations under uncertainty.}




\maketitle

\input{sections/intro/intro}
\input{sections/background/background}

\input{sections/contribution/DTCalib}
\input{sections/contribution/GFlownet}
\input{sections/experiments/validation}
\input{sections/background/related}
\input{sections/conclusion/conclusion}

\bibliography{sn-bibliography}

\end{document}

%% file: preamble.tex
\usepackage{graphicx}%
\usepackage{multirow}%
\usepackage{amsmath,amssymb,amsfonts}%
\usepackage{amsthm}%
\usepackage{mathrsfs}%
\usepackage[title]{appendix}%
\usepackage{xcolor}%
\usepackage{textcomp}%
\usepackage{manyfoot}%
\usepackage{booktabs}%
\usepackage{algorithm}%
\usepackage{algorithmicx}%
\usepackage{algpseudocode}%
\usepackage{listings}%
\usepackage{ifthen}
\usepackage{multirow}
\usepackage{booktabs}
\usepackage{float}
\usepackage{siunitx}
\usepackage{tabularx}
\newboolean{showcomments}
\setboolean{showcomments}{true} 
\ifthenelse{\boolean{showcomments}}
{\newcommand{\nb}[2]{
    \fcolorbox{gray}{yellow}{\bfseries\sffamily\scriptsize#1}
    {$\blacktriangleright$#2$\blacktriangleleft$}
  }
  
}
{\newcommand{\nb}[2]{}
  
}

\newcommand{\Fig}[1]{Fig.~\ref{#1}}  			

\newcommand{\rone}{\textbf{R1}}
\newcommand{\rtwo}{\textbf{R2}}

%% file: sections/intro/intro.tex
\section{Introduction}
\label{sec:intro}
Digital twins have become an important technology for the monitoring, prediction, and control of complex systems, particularly in industrial and cyber-physical settings~\cite{Zhou2022Revisiting,Javaid2023DTIndustry40}. More recently, the concept has also been extended to natural systems in domains such as smart farming~\cite{david2023digital} and health sciences~\cite{katsoulakis2024digital}. This extension raises distinct modeling challenges. In natural systems, observations are often sparse, noisy, delayed, destructive, or costly to acquire, while uncertainty remains central throughout model updating and downstream use~\cite{lin2021uncertainty}.

In such settings, digital twin services rely on mechanistic simulators that encode domain knowledge about the underlying physical, chemical, or biological processes. Their usefulness depends on whether they remain consistent with the physical system being twinned. This raises a model-adaptation problem: simulator parameterizations must be updated so that simulated behavior remains credible in the target context, for example under local environmental conditions, cultivar-dependent traits, or configuration-specific responses. Because the objective is to infer parameter settings from observed outputs rather than predict outputs from known parameters, this is naturally an inverse problem. In digital twins of natural systems, however, sparse and indirect observations often fail to identify a unique calibration. Several parameterizations may remain compatible with the same observed behavior, especially when many parameters are not directly measurable~\cite{gallo2022lack}.

This motivates the use of simulation-based inference, which addresses parameter inference in simulator-based models when the likelihood is unavailable or intractable. Candidate parameterizations can be proposed, instantiated in the simulator, and evaluated through their discrepancy with observed trajectories without requiring an explicit analytical likelihood. Within this setting, Generative Flow Networks~\cite{bengio2023foundations} (GFlowNets) are appealing because they learn reusable stochastic policies that generate complete configurations with probability proportional to a reward. This makes them well suited to adaptation settings in which several high-quality parameterizations may remain plausible and should be preserved explicitly rather than collapsed into a single fitted solution.

This article studies GFlowNet-based model adaptation in a digital twin setting for greenhouse crop modeling, where a mechanistic tomato growth simulator~\cite{vanthoor2011methodology} is adapted to new observational contexts from sparse trajectories. The greenhouse setting provides a useful testbed because it allows the approach to be analyzed first in a fully enumerable regime, where all candidate parameterizations in the considered search space can be evaluated exactly, and then in a larger regime where exhaustive characterization is no longer feasible. The main contributions of this work are as follows:

\begin{itemize}
    \item We formulate mechanistic model adaptation in digital twins of natural systems as a simulation-based inference problem, and position it as a dedicated service within the digital twin lifecycle under sparse observations and structural non-identifiability.
    \item We introduce a generative adaptation mechanism based on GFlowNets that preserves and samples multiple plausible simulator parameterizations, rather than reducing adaptation to a single fitted calibration.
    \item We report an empirical study on a smart farming model, examining candidate retrieval, alignment with an enumerable target landscape, and the behavior of the approach as the adaptation space becomes larger and less tractable.
\end{itemize}

The remainder of this article is organized as follows. Section~\ref{sec:background} reviews digital twins of natural systems and GFlowNets. Section~\ref{sec:sbi} formulates model adaptation as a simulation-based service within the digital twin lifecycle and introduces the requirements that guide our approach. Section~\ref{sec:gfn} presents the proposed generative formulation and its GFlowNet-based implementation. Section~\ref{sec:eval} showcases our results and discusses their implications for the model adaptation service, Section~\ref{sec:related} presents the related works, and Section~\ref{sec:conclusion} concludes the paper.

%% file: sections/background/background.tex
\section{Background}
\label{sec:background}
This section reviews the two main concepts framing our approach. We first introduce digital twins of natural systems and the particular challenges they raise for model adaptation under uncertainty. We then present GFlowNets as generative models for structured spaces.

\subsection{Digital twins of natural systems}
\label{sec:back-dt}
Digital twins are virtual representations of physical systems that remain linked to their real-world counterparts through data exchange and model updating, enabling monitoring, prediction, and decision support~\cite{kapteyn2021probabilistic}. Recent work in model-driven engineering further distinguishes digital twins by the nature of the system being twinned, separating engineered from non-engineered contexts and explicitly including biological systems within the latter~\cite{michael2025mde}. This distinction matters because moving from artifacts whose structure is largely known by design to natural systems introduces forms of variability, partial observability, and uncertainty that are not controlled in the same way. Natural systems may evolve through phases such as plan, growth, life, and end-of-life~\cite{michael2025mde}. Across these phases, the digital twin must coordinate model artifacts, contextual data, and downstream services while the underlying biological or environmental system continues to change. Unlike many engineered systems whose relevant structure is largely fixed by design, natural systems may vary substantially over time as a result of development, environmental exposure, or management conditions. The models embedded in the twin then cannot be treated as static artifacts calibrated once and reused unchanged. They must remain suitable for the current lifecycle phase and operating context of the physical system.

Digital twins for natural systems have recently gained attention in domains such as ecology~\cite{de2023digital}, and medicine~\cite{an2022drug}. In smart farming, they are commonly associated with the integration of sensing, simulation, and data-driven services to support monitoring, what-if analysis, and operational decision-making across greenhouses, fields, and livestock systems~\cite{pylianidis2021agriculture}. At the same time, the literature suggests that these systems remain comparatively immature, and that robust predictive and prescriptive capabilities are still limited in many practical deployments~\cite{purcell2023agriculture}.

Maintaining the twin involves more than synchronizing observations with a virtual replica, it also requires updating the underlying models so that they remain suitable for the current environmental context. This is especially difficult in natural systems, where observations are often partial, noisy, delayed, destructive, or costly to obtain, and where uncertainty propagates through both model updating and downstream decision-making~\cite{thelen2023comprehensive}. As a result, several model parameterizations may remain consistent with the available evidence. Model adaptation is therefore more naturally viewed as an inference problem over a set of plausible configurations than as a straightforward search for a single fixed calibration.

\subsection{GFlowNets}
\label{sec:back-gfn}

GFlowNets~\cite{bengio2021flow} are generative models designed for problems in which complete solutions are constructed through a sequence of dependent decisions and where several distinct high-quality solutions may all be relevant. Given a positive reward $R(x)$ defined over complete solutions $x$, a trained GFlowNet samples terminal objects with probability proportional to a reward function, so that higher-reward solutions are sampled more frequently while lower-reward alternatives retain proportionally smaller but non-zero mass. They are therefore well suited to multimodal inference settings in which the objective is not only to identify one good candidate, but also to represent and sample multiple plausible ones.

Formally, a GFlowNet operates on a directed acyclic graph whose nodes represent states, corresponding to partial objects under construction, and whose edges represent actions that extend one state into another. An initial state encodes an empty or default configuration, while terminal states correspond to complete candidate solutions. A GFlowNet learns a forward policy over this construction graph such that terminal objects are sampled in proportion to the reward. That is, for policy $\pi(x)$:

\begin{equation}
    \pi(x) \propto R(x)
\end{equation}

The term \emph{flow} refers to the propagation of probability mass through the graph from the initial state toward terminal states, with training encouraging local transition probabilities to induce this reward-proportional distribution over complete objects~\cite{bengio2023foundations}.

This reward-proportional sampling distinguishes GFlowNets from methods designed to return a single optimum, such as gradient-based optimization or reward-maximizing reinforcement learning. Rather than concentrating search around one maximizer, GFlowNets learn a reusable policy whose sampling frequencies are calibrated to reward magnitudes across the entire solution space. This makes them useful both in multimodal search problems and in settings where samples must be generated repeatedly without solving a new search problem from scratch each time. Such uses have been reported, for example, in multi-objective optimization~\cite{jain2023multiobj}, posterior approximation over graph structures in Bayesian structure learning~\cite{deleu2022bayesian}, and amortized inference for large language models~\cite{hu2023amortizing}. These properties are directly relevant to simulator adaptation. When observations are sparse and several parameterizations can reproduce the observed trajectories with similar fidelity, the objective is not only to identify a strong candidate, but also to represent and sample multiple plausible ones. In this sense, GFlowNets provide a natural bridge toward the generative adaptation perspective developed in Section~\ref{sec:gfn}.

%% file: sections/contribution/DTCalib.tex
\section{Simulation-based model adaptation in digital twins of natural systems}
\label{sec:sbi}
This section positions model adaptation as a service within the lifecycle of a digital twin of a natural system. We first situate this service with respect to the model artifacts and downstream services that depend on it. We then describe the internal simulation-based workflow through which candidate parameterizations are generated, evaluated, and refined from observations collected in the target context. On this basis, we finally identify the requirements that motivate the generative approach developed in Section~\ref{sec:gfn}.

\subsection{Model adaptation in the digital twin lifecycle}
\label{sec:adaptation-dt}
Because digital twins of natural systems are tied to physical systems that evolve through phases such as plan, growth, life, and end-of-life~\cite{michael2025mde}, they must coordinate model artifacts and services across that lifecycle. These artifacts include model specifications, mechanistic simulators, configuration data, and contextual observations collected from the physical system. On top of them, the twin supports services such as monitoring, forecasting, what-if analysis, synchronization, and operational decision support. In the setting considered here, the mechanistic simulator is a central artifact because several downstream services depend on it as their behavioral model. As the natural system evolves, the twin must maintain coherence between the current observations, the simulator configuration, and the services that consume it. This is where model adaptation becomes necessary: not as an isolated calibration exercise, but as the mechanism through which the simulator remains suitable for the current context and lifecycle phase of the physical system.

Accordingly, model adaptation is best treated as a service within the digital twin lifecycle rather than as a calibration step performed only once during initialization. Its role is to take a base simulator together with contextual observations and produce an adapted model configuration that can be consumed by downstream digital twin services. In practice, this service maintains the operational relevance of the mechanistic simulator as new evidence becomes available, operating conditions change, or the natural system moves from one lifecycle phase to another.

\begin{figure}
  \centering
  \includegraphics[width=\textwidth]{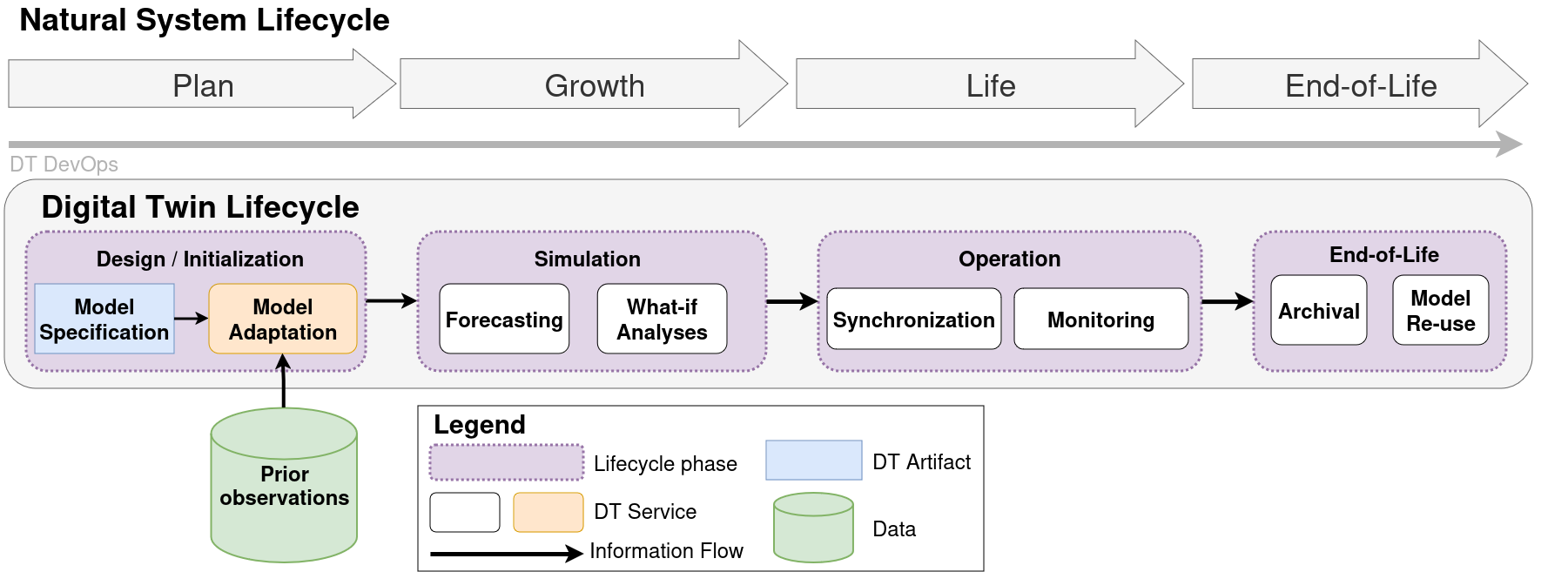}
  \caption{Model adaptation as a service within the lifecycle of a digital twin for a natural system. During design and initialization, the service combines a model specification with contextual observations to produce an adapted simulator that can be consumed by downstream simulation and operational services.}
  \label{fig:sbi-lifecycle}
\end{figure}

\Fig{fig:sbi-lifecycle} situates this service within the design and initialization phase of the digital twin lifecycle, where a model specification and observations collected in the target context are combined to produce an adapted model. This adapted model is then passed to simulation services, including forecasting and what-if analysis, and ultimately supports downstream operational activities. The figure emphasizes that adaptation is not an isolated preprocessing step, but a model-management service that sustains the link between simulator specification, simulation, and use.

This role is especially important in natural systems, where the information required to update the simulator is often incomplete~\cite{blair2021digital}. Many simulator parameters cannot be measured directly, while observations may be sparse, noisy, delayed, destructive, or costly to obtain. Moreover, similar observed trajectories may remain consistent with several simulator parameterizations. For this reason, model adaptation is not necessarily invoked continuously in real time. It may instead be triggered when new observations become available, when the physical system enters a new lifecycle phase, or when discrepancies between simulated and observed behavior indicate that the current simulator is no longer adequate for the target context. Model adaptation is thus a recurring service that supports the continued use of the simulator across the lifecycle of the natural system.

\subsection{Simulation-based model adaptation}
\label{sec:adaptation-sim}

Having situated model adaptation within the digital twin lifecycle, we now turn to the internal workflow through which this service operates. In the present setting, adaptation is formulated as a simulation-based inference problem. Starting from a base mechanistic simulator and observations collected in a target context, the service seeks parameterizations whose simulated behavior remains consistent with the available evidence. Here, the target context denotes the conditions under which the physical system is observed and the simulator is intended to be used, for example a given growth stage, environmental regime, or management setting.

\begin{figure}
  \centering
  \includegraphics[width=\textwidth]{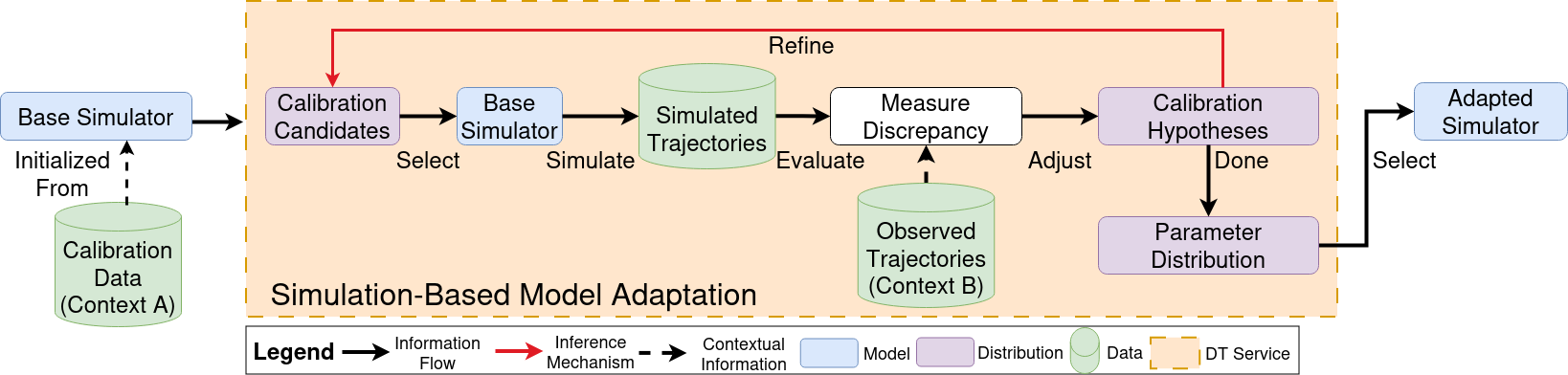}
    \caption{Simulation-based workflow of the model adaptation service. Starting from a base simulator and observations collected in the target context, the service iteratively generates candidate parameterizations, executes the simulator, compares simulated and observed trajectories, and refines a distribution over plausible calibration hypotheses. The resulting adapted simulator, or set of plausible adapted configurations, can then be consumed by downstream digital twin services.}
  \label{fig:sbi-flow}
\end{figure}

\Fig{fig:sbi-flow} summarizes this workflow. The process begins from a base simulator that has been calibrated on an original context. Candidate parameterizations are then instantiated by the service and are then simulated to produce trajectories under the target context. These simulated trajectories are compared with the observed trajectories through a discrepancy measure that reflects adaptation quality. The resulting scores are used to update a distribution over calibration hypotheses and to guide the generation of subsequent candidates. Through repeated simulation and evaluation, the service refines the set of plausible parameterizations and returns an adapted simulator configuration, or more generally a set of plausible adapted configurations, for downstream use.

This workflow is simulation-based because candidate hypotheses are assessed through forward execution of the mechanistic simulator rather than through an explicit analytical likelihood model~\cite{cranmer2020frontier}. This is well suited to the types of simulators considered here, where the relevant quantities can be generated by execution but the inverse mapping from observations to parameters is unavailable in closed form. Here, the inference target is the simulator parameterization rather than the time-evolving hidden state of a fixed model. The objective is to identify mechanistic parameter settings that make the simulated trajectories consistent with the observed system behavior with sufficient fidelity for downstream digital twin services.

This formulation highlights two consequences that motivate the remainder of the paper. First, because observations constrain parameters only indirectly through simulated behavior, adaptation must proceed through repeated hypothesis generation and evaluation. Second, because the available evidence may remain insufficient to isolate a unique calibration, the service must reason over a set of plausible configurations rather than over a single fitted parameter vector. These observations motivate the requirements introduced next.

\subsection{Requirements of the model adaptation service}
\label{sec:requirements}
Model adaptation in digital twins of natural systems must proceed from partial and indirect observations of the physical system. The data available for adaptation do not consist of direct measurements of simulator parameters. Instead, they take the form of observed outputs or trajectories, such as harvest, biomass accumulation, or other aggregate metrics shaped by the joint action of multiple system configurations over time. As these measurements reflect the combined effects of several interacting parameters, they provide only indirect evidence about which internal simulator configurations remain plausible. For example, a crop exposed to warmer daytime temperatures and cooler nights may reach a similar observed harvest outcome as one exposed to a milder but more stable temperature regime, even though the underlying trajectories differ. Thus, the adaptation service must infer which parameterizations remain consistent with the observed system behavior, rather than recover parameter values directly from the data. This inverse relation is often underdetermined: several distinct simulator configurations may remain compatible with the same evidence, especially when observations are sparse, aggregated, or restricted to a subset of system variables. We identify two requirements that the model adaptation service must satisfy.

\textbf{Requirement 1 (R1): Simulation-based model adaptation.}
During the design phase of a digital twin of a natural system, model adaptation must remain grounded in the mechanistic simulator that will support downstream digital twin services. At this stage, the objective is not only to improve agreement with available observations, but to establish a simulator configuration whose parameters remain consistent with the model and whose behavior remains credible for subsequent use in forecasting, what-if analysis, and decision support. This is particularly important in natural-system models, where the effects of many parameters are correlated and expressed through only partially observed dynamics. The contribution of any single parameter may then be difficult to isolate, since measured outputs reflect the combined action of several interacting processes over time. Under these conditions, direct statistical inference of model parameters is unsuitable for adaptation. The service must instead update its beliefs over plausible parameterizations through likelihood-free methods such as simulation-based inference. Because these parameterizations are evaluated through forward simulation of a mechanistic model, they preserve the interpretability and internal consistency required for use in other digital twin services, while keeping the model aligned with the twinned natural system.

\textbf{Requirement 2 (R2): Preservation of plausible parameterizations.}
Model adaptation in natural systems must avoid collapsing unresolved ambiguity into a single calibrated solution. A single retained calibration may appear well supported while masking other parameterizations that are equally consistent with the observed system behavior, and better reflect the dynamics of the natural twin. Thus, the objective of the adaptation service should not be limited to selecting one best-fitting configuration, but should aim to preserve multiple plausible simulator parameterizations that remain compatible with the target context. Making an early commitment to one model configuration would hide epistemic uncertainty that has not been resolved by the evidence. In a digital twin, that uncertainty matters for downstream services, since forecasting, what-if analysis, and decision support may depend on which plausible parameterization is retained. Preserving several compatible parameterizations allows the twin to quantify uncertainty over model adaptation and make it available to downstream services, whose sensitivity to that uncertainty may differ across monitoring, predictive, and prescriptive tasks.

Taken together, these requirements define adaptation not as the recovery of a single calibrated parameter vector, but as the identification and maintenance of plausible simulator configurations supported by observed system behavior. This view motivates the generative approach developed in the next section.

%% file: sections/contribution/GFlownet.tex
\section{GFlowNet-based model adaptation}
\label{sec:gfn}
The previous section established two requirements for model adaptation in digital twins of natural systems: adaptation must remain grounded in the mechanistic simulator, and it must preserve multiple plausible parameterizations when the available evidence does not justify a unique calibration. We address these requirements by casting model adaptation as a generative modeling problem over simulator configurations. Section~\ref{sec:generative-adaptation} develops this perspective, Section~\ref{sec:gfn-inference} specializes it to GFlowNets, and Section~\ref{sec:training} presents the corresponding training and inference procedure.

\subsection{Generative modeling for model adaptation}
\label{sec:generative-adaptation}
Standard simulation-based inference methods are often framed as posterior-estimation problems: given observations, the objective is to recover a distribution over parameters that is consistent with the simulator and the data. In the present setting, however, the adaptation service must do more than characterize which parameter regions are plausible. It must also produce complete simulator configurations that can be simulated, compared, and subsequently used by downstream digital twin services. This becomes particularly important when the relation between observed behavior and simulator parameters is underdetermined, so that several disconnected or structured regions of parameter space may remain plausible.

A generative modeling perspective addresses this need by shifting the objective from posterior recovery to parameter generation. Here, generative modeling refers to learning a model that can construct complete simulator configurations and sample them from a distribution over plausible parameterizations. The mechanistic simulator remains the object being adapted, and candidate configurations are still evaluated through their simulated agreement with observations. The generative model therefore does not replace the mechanistic structure with a purely statistical surrogate. Its role is to represent and sample plausible simulator configurations in a form that preserves uncertainty over adaptation while remaining directly usable for downstream digital twin services.

Among generative models, GFlowNets are particularly well suited to the present problem because they learn to construct structured objects while sampling them approximately in proportion to a reward function. In the present setting, that reward is derived from agreement between simulated and observed system behavior, which makes GFlowNets a natural candidate for representing and exploring the space of plausible simulator parameterizations.

\subsection{GFlowNet model for parameter inference}
\label{sec:gfn-inference}
Our GFlowNet defines a stochastic policy over the discrete space of candidate simulator configurations, so that complete parameterizations are generated with probability proportional to a positive reward. In the present setting, terminal states correspond to complete simulator parameterizations, and the learned policy acts as a generative adaptation mechanism over mechanistic model configurations. The resulting object is not a single calibrated parameter vector, but a sampler over plausible simulator parameterizations whose probabilities are shaped by agreement with observed system behavior.

Let $\mathcal{X}$ denote the discrete space of simulator configurations induced by the perturbation scheme used for model adaptation. Each terminal state $x \in \mathcal{X}$ corresponds to a complete parameterization $\boldsymbol{\theta}(x)$ obtained from a reference configuration $\boldsymbol{\theta}_0$ through a sequence of bounded perturbations. A non-terminal state $s_t$ represents a partial construction after $t$ decisions, and a trajectory
\begin{equation}
\label{eq:gfn-trajectory}
\tau=(s_0,a_1,s_1,\dots,a_T,s_T=x)
\end{equation}
corresponds to the sequential construction of one full simulator configuration. In the setting considered here, actions are defined over mutually exclusive parameter groups, and each action applies a discrete perturbation to one group. Parameter inference is thus formulated as a structured parameter-set construction problem.

\begin{figure}[H]
    \centering
    \includegraphics[width=\textwidth]{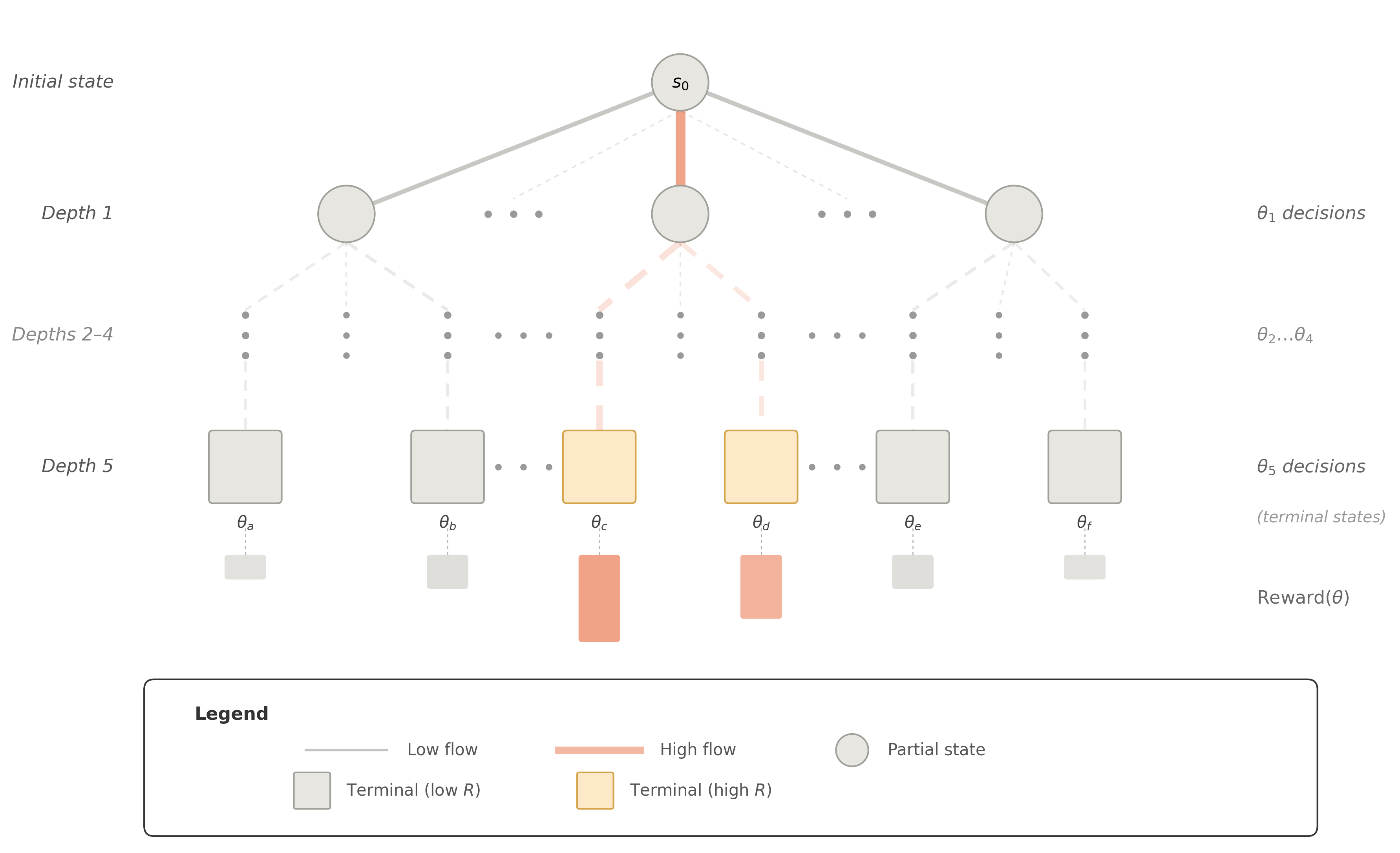}
    \caption{Structured learnable space induced by the GFlowNet adaptation process. Starting from an initial state, the policy constructs a complete simulator parameterization through sequential decisions over parameter groups. Intermediate nodes represent partial configurations, and terminal nodes represent complete parameterizations.}
    \label{fig:gfn}
\end{figure}

The learnable space induced by this construction is illustrated in~\Fig{fig:gfn}. Rather than selecting a complete simulator parameterization in a single step, the GFlowNet constructs it sequentially through a series of decisions over parameter groups. Partial states correspond to intermediate configurations in which only a subset of decisions has been made, whereas terminal states correspond to complete simulator parameterizations. This sequential construction defines the space explored by the adaptation policy. During training, probability mass is progressively concentrated along branches that lead to higher-reward parameterizations, while alternative branches remain accessible when they also correspond to plausible mechanistic configurations.

Let $\mathcal{C}=\{1,\dots,C\}$ denote the set of observational contexts used for adaptation, and let $\mathbf{y}^{\mathrm{obs}}_c$ denote the observations associated with context $c \in \mathcal{C}$. For a candidate configuration $x$, the mechanistic simulator $\mathcal{M}$ produces
\begin{equation}
\label{eq:simulator-context}
\mathcal{M}(\boldsymbol{\theta}(x),c)\mapsto \mathbf{y}^{\mathrm{sim}}_c(x)
\end{equation}

More precisely, let $\theta_t$ denote the current parameterization after $t$ group\mbox{-}level decisions, with initial state $\theta_0$ equal to the reference configuration. For a decision taken at cycle $c$ on group $g_t$ with action $a_t$, each affected parameter $p$ is updated according to
\begin{equation}
\theta_{t+1,p}
=
\operatorname{clip}\!\left(
\theta_{t,p}
+
\eta_c\ \cdot sf\ \cdot \operatorname{sign}_{a_{t},p}(x)\cdot (u_p - l_p),
\, l_p,\, u_p
\right).
\end{equation}
where $sf \in (0,1]$ is the step fraction, $\operatorname{sign}_{a_{t},p}(x)\in\{-1,0,+1\}$ encodes the direction of the perturbation prescribed by the selected group--action pair for parameter $p$, and $(l_p, u_p)$ are the admissible bounds of parameter $p$. The operator $\operatorname{clip}(\cdot,l_p,u_p)$ ensures that updated values remain within the valid parameter range. Parameters for which $\operatorname{sign}_p(x)=0$ remain unchanged. In the multi\mbox{-}cycle setting, the effective update magnitude is annealed across cycles through $\eta_c = 2^{-(c-1)}$, so that later passes refine earlier decisions more conservatively. Repeated application of this update rule over the ordered parameter groups yields the complete simulator parameterization $\theta(x)$ associated with terminal state $x$.

Adaptation quality is evaluated jointly across contexts rather than from a single trajectory. Thus, we define the contextual discrepancy
\begin{equation}
\label{eq:context-discrepancy}
\ell_c(x)=D\!\left(\mathbf{y}^{\mathrm{sim}}_c(x),\mathbf{y}^{\mathrm{obs}}_c\right)
\end{equation}
where $D(\cdot,\cdot)$ denotes the trajectory-level discrepancy used for calibration. In practice, $D$ is computed from normalized residuals so that variables or time points with larger magnitudes do not dominate the adaptation criterion purely because of scale. The resulting contextual loss vector is
\begin{equation}
\label{eq:context-loss-vector}
\boldsymbol{\ell}(x)=\bigl(\ell_1(x),\dots,\ell_C(x)\bigr)
\end{equation}
which summarizes how well one simulator configuration agrees with the observations in each context.

A difficulty in this setting is that raw contextual losses can be highly heterogeneous. Some contexts are systematically easier to fit than others, and the scale of the discrepancy may vary substantially across contexts and across the explored state space. If used directly, these differences can make the reward either too flat or overly concentrated, which weakens the ability of the GFlowNet to preserve meaningful diversity. To regularize this effect, each contextual loss is normalized through empirical quantiles:
\begin{equation}
\label{eq:quantile-normalized-loss}
\widetilde{\ell}_c(x)=
\frac{\ell_c(x)-q^{(c)}_{\mathrm{lo}}}
     {q^{(c)}_{\mathrm{hi}}-q^{(c)}_{\mathrm{lo}}+\varepsilon}
\end{equation}
where $q^{(c)}_{\mathrm{lo}}$ and $q^{(c)}_{\mathrm{hi}}$ denote lower and upper quantiles estimated from the distribution of losses in context $c$, and $\varepsilon>0$ avoids numerical instability. The normalization in Eq.~\eqref{eq:quantile-normalized-loss} makes contextual losses more comparable and makes reward shaping less sensitive to context-specific scale effects.

Robustness across contexts is then incorporated directly into the adaptation criterion. Rather than aggregating the normalized losses in Eq.~\eqref{eq:quantile-normalized-loss} by a simple mean alone, we sort them in decreasing order,
\begin{equation}
\label{eq:sorted-context-loss}
\widetilde{\ell}_{(1)}(x)\geq \widetilde{\ell}_{(2)}(x)\geq \dots \geq \widetilde{\ell}_{(C)}(x)
\end{equation}
and define a tail-sensitive score over the $K$ worst contexts:
\begin{equation}
\label{eq:tail-loss}
\mathcal{L}_{\mathrm{tail}}(x)=\frac{1}{K}\sum_{j=1}^{K}\widetilde{\ell}_{(j)}(x)
\end{equation}
The overall adaptation loss is then written as
\begin{equation}
\label{eq:adaptation-loss}
\mathcal{L}(x)=
(1-\lambda)\frac{1}{C}\sum_{c=1}^{C}\widetilde{\ell}_c(x)
+\lambda\,\mathcal{L}_{\mathrm{tail}}(x)
\end{equation}
where $\lambda \in [0,1]$ controls the importance of robustness to hard contexts. 

Eq.~\eqref{eq:adaptation-loss} makes the adaptation objective tail-risk-sensitive. That is, parameterizations are favored not only for good average fit, but also for avoiding poor performance on the worst-performing contexts~\cite{rockafellar2000optimization}. This is useful here because digital twin models may need parameterizations that remain credible across several operating conditions rather than configurations that perform well only on average. This makes model adaptation more robust to parameterizations that perform poorly on a subset of contexts.

Following the standard GFlowNet formulation, the terminal reward defines a positive unnormalized target distribution over complete parameterizations. We instantiate this target in Boltzmann form from the contextual loss in Eq.~\eqref{eq:adaptation-loss}:
\begin{equation}
\label{eq:boltzmann-reward}
R(x)=\exp\!\bigl(-\beta \,\mathcal{L}(x)\bigr)
\end{equation}
so that parameterizations with lower risk receive exponentially higher reward, and $\beta>0$ controls how strongly probability mass is concentrated around them.

Under this construction, the GFlowNet learns to sample complete parameterizations with frequency proportional to the reward in Eq.~\eqref{eq:boltzmann-reward}. This proportionality biases sampling toward the best-performing regions of the reward landscape, while still assigning probability mass to alternative plausible configurations according to their relative quality.

This formulation differs from point calibration in a crucial way: the objective is not to identify one parameterization that minimizes a global discrepancy, but to learn a generative policy over complete mechanistic configurations whose sampling probabilities reflect both fit and robustness across contexts. In this way, the GFlowNet satisfies \rone{} by keeping inference grounded in the mechanistic simulator, while operationalizing \rtwo{} through explicit generation of multiple plausible simulator configurations.

\subsection{Training and inference procedure}
\label{sec:training}
Training proceeds by repeatedly sampling trajectories in the GFlowNet state space, evaluating the terminal parameterizations reached by those trajectories, and updating the forward policy so that terminal sampling frequencies align with the reward-defined target distribution. For a terminal state $x$, this requires simulating the mechanistic model across the observational contexts in Eq.~\eqref{eq:simulator-context}, computing the contextual discrepancies in Eq.~\eqref{eq:context-discrepancy}, normalizing them according to Eq.~\eqref{eq:quantile-normalized-loss}, and aggregating them into the tail-risk-sensitive adaptation loss defined in Eqs.~\eqref{eq:tail-loss} and~\eqref{eq:adaptation-loss}. The resulting score is then transformed into the terminal reward through the Boltzmann construction in Eq.~\eqref{eq:boltzmann-reward}. Simulator execution is the most expensive component of the training pipeline, since it provides the trajectory-level evidence from which terminal reward is constructed.

Depending on the size of the terminal state space, reward construction can be handled in two ways. When the set of complete parameterizations is small enough to enumerate, simulator outputs, contextual losses, and terminal rewards can be computed offline and cached for all terminal states. In larger settings, these quantities are obtained on demand when terminal states are encountered during training. In both cases, the optimization problem remains the same: the GFlowNet is trained so that its terminal distribution matches the reward-induced target distribution over complete simulator configurations.

Let $P_F(\tau)$ and $P_B(\tau)$ denote the forward and backward trajectory probabilities associated with a complete construction trajectory $\tau$ in Eq.~\eqref{eq:gfn-trajectory}, ending in terminal state $x$. The model is trained with the trajectory balance objective~\cite{malkin2022trajectory}, which relates trajectory probabilities and terminal rewards through
\begin{equation}
\label{eq:trajectory-balance}
\log Z + \log P_F(\tau) - \log P_B(\tau) \approx \log R(x),
\end{equation}
where $Z$ is the normalizing constant of the target distribution. In practice, training minimizes the squared trajectory balance residual associated with Eq.~\eqref{eq:trajectory-balance} over sampled trajectories. In the discrete perturbation setting considered here, the construction graph is a tree, since each complete parameterization is reached by a unique sequence of decisions. Each non-initial state has a single parent, which makes the backward policy deterministic. As a result, $P_B(\tau)$ is fixed and resolves to one, thus no learned backward policy is required. The training signal is  determined solely by the forward construction policy, the terminal reward in Eq.~\eqref{eq:boltzmann-reward}, and the learned normalizing constant.

Operationally, this separates offline adaptation from deployment-time sampling. During the design or update phase, observations from the target context are used to construct the reward landscape and train the GFlowNet, so that the mechanistic simulator remains consistent with the target system in accordance with \rone{}. Once training is complete, the learned policy can be queried at low cost to generate multiple plausible parameterizations on demand, which supports \rtwo{} by making uncertainty over model adaptation available to downstream digital twin services without repeating the full adaptation procedure.


%% file: sections/experiments/validation.tex
\section{Evaluation}
\label{sec:eval}
To assess the proposed generative model adaptation mechanism, the experimental study is organized around the following research questions:

\begin{itemize}
    \item \textbf{RQ1 - Mode Discovery}: to what extent can the proposed GFlowNet recover distinct high-quality regions of the adaptation landscape, thereby enabling the generation of diverse and plausible simulator parameterizations under sparse observations?
    \item \textbf{RQ2 - Retrieval Performance}: how effectively does the proposed approach retrieve strong adaptation hypotheses, and how does it compare with baseline search methods under matched simulator-evaluation budgets?
    \item \textbf{RQ3 - Scalability}: how does the proposed approach behave as the adaptation problem moves from an enumerable regime to larger and more computationally demanding settings, and what practical limitations emerge for real-system model adaptation?
\end{itemize}

The rest of this section is organized as follows: Section~\ref{sec:case-study} presents the case studied in our experiments, Section~\ref{sec:experiments} details our experimental setup, Section~\ref{sec:results} showcases our results, Section~\ref{sec:discussion} discusses the relevance of our approach to the requirements presented in Section~\ref{sec:requirements}, and Section~\ref{sec:threat} describes threats to validity.

\input{sections/experiments/casestudy}
\input{sections/experiments/experiments}
\input{sections/experiments/results}
\input{sections/experiments/discussion}
\input{sections/experiments/threats}

%% file: sections/experiments/casestudy.tex
\subsection{Case Study}
\label{sec:case-study}
Controlled environment agriculture refers to crop production systems in which key environmental variables such as temperature, light, humidity, CO$_2$, irrigation, and nutrient supply are actively regulated in order to shape plant development and resource use. Greenhouse horticulture is one of the most established forms of Controlled environment agriculture, and it provides a particularly relevant setting for model-based reasoning because crop performance emerges from the interaction between biological processes and tightly managed climate-control strategies. Controlled environment agriculture has also been recognized as a relevant setting for digital twin development, where models, observations, and decision-support services must be combined in a coherent operational framework~\cite{eramo2022conceptualizing}. These characteristics make greenhouse production an informative application domain for the present work: it is mechanistically rich, operationally data-intensive, and representative of the broader class of natural systems for which digital twins must integrate heterogeneous models and observations~\cite{david2023digital}. Recent reviews further suggest that smart farming remains a useful testbed for studying such model-based digital twin mechanisms in practice~\cite{subeesh2025agricultural}.

The empirical study is based on data from the second edition of the Autonomous Greenhouse Challenge, conducted in six high-tech glasshouse compartments at the Wageningen Research Centre in Bleiswijk, The Netherlands~\cite{hemming2020AGC2}. The production cycle spans approximately six months and concerns cherry tomato cultivation of cultivar \emph{Axiany}. Five compartments were operated remotely by multidisciplinary international teams, while a sixth compartment was managed manually by commercial Dutch growers and used as a reference. The objective of the challenge was to maximize net profit by balancing production and fruit quality against resource consumption, including water, nutrients, heating, electricity, and CO$_2$. For the purposes of the present study, we use the subset of the challenge data that is relevant to simulator adaptation, namely production data and environmental measurements collected under the different greenhouse control strategies.

This dataset is well suited to the present study because it exposes the adaptation mechanism to several observational contexts generated under different control policies. From the perspective of simulator calibration, each compartment provides a distinct realization of the same crop-production process under a different management regime. The inference problem is therefore not to match a single trajectory in isolation, but to identify simulator parameterizations that remain compatible with multiple observed trajectories collected under heterogeneous operating conditions. This setting is closely aligned with the motivation of the paper: sparse and noisy observations may support several plausible mechanistic explanations, so a useful adaptation mechanism should support not only good fit, but also the recovery of multiple strong hypotheses.

The mechanistic simulator used in our study is derived from the tomato yield model of Vanthoor et al., originally developed for model-based greenhouse design~\cite{vanthoor2011methodology}. The model describes tomato production as a function of indoor climate, with particular emphasis on the effects of temperature, light, and CO$_2$ concentration on yield. Its internal structure is organized around a carbohydrate buffer that mediates assimilate allocation to fruits, leaves, and stems, while temperature-dependent growth inhibition functions account for both instantaneous and mean temperature effects on crop production. This makes it a suitable mechanistic basis for the present experiments: the model remains physiologically interpretable, yet can be simulated repeatedly under many candidate parameterizations.

%% file: sections/experiments/experiments.tex
\subsection{Experimental Setup}
\label{sec:experiments}
This subsection describes the calibration problem, the compared methods, and the evaluation protocol used throughout the empirical study. The experiments are designed to assess the three research questions introduced above, namely mode discovery, retrieval performance under matched simulator-evaluation budgets, and scalability as the adaptation space becomes too large for exhaustive analysis.

\subsubsection{Crop Simulator and Observation Protocol}
\label{sec:experiments-protocol}
In the present study, adaptation is carried out over six observational contexts, each corresponding to one greenhouse compartment from the dataset introduced in the case study. For each candidate parameterization, the simulator is executed independently on all six observed contexts. The resulting simulated trajectories are compared with the corresponding observations. Rather than matching the full simulated time series, the adaptation signal is constructed from a finite set of observation points extracted from each trajectory. These points are concentrated near the end of the simulation horizon, where accumulated differences between parameterizations become most informative for the crop-production variables of interest. In practice, the observations correspond to crop yield dry mass measured according to the greenhouse collection schedule, namely once every two weeks.

The six context-specific discrepancies are then combined into the contextual loss defined in Section~\ref{sec:gfn-inference}. A parameterization is considered good only to the extent that it provides coherent agreement across the full observation set rather than fitting one context in isolation. This makes the adaptation problem more constrained than single-trajectory fitting while preserving the possibility that several distinct parameterizations remain plausible under sparse observations. The rationale is that adapted simulator parameters should capture crop-growth variation across contexts rather than overfit a single greenhouse trajectory.

\subsubsection{Parameter Grouping and Search-Space Construction}
Calibration is performed by perturbing a baseline simulator parameterization rather than by estimating all simulator parameters freely and independently. The chosen baseline parameterization corresponds to the parameters derived by Vanthoor et al.~\cite{vanthoor2011methodology} for their model. The adjustable parameters are organized into five mutually exclusive groups obtained from a physically motivated functional decomposition. More specifically, this decomposition follows a physical interpretation of the growth process encoded by the simulator: incoming radiation first interacts with the canopy, which affects canopy temperature, absorption, and subsequent physiological processes that ultimately determine growth and yield. The groups are defined so that each one corresponds to a coherent component of this progression rather than to an arbitrary subset of coordinates. As a result, perturbations correspond to coherent model modifications rather than arbitrary coordinate changes. Table~\ref{tab:vanthoor_parameter_groups} summarizes this grouping.

\begin{table*}[t]
\centering
\footnotesize
\begin{tabularx}{\textwidth}{p{0.05\textwidth}p{0.18\textwidth}p{0.06\textwidth}p{0.22\textwidth}X}
\toprule
Order & Group & \# Params. & Possible perturbations & Vanthoor model parameters \\
\midrule
1 & Leaf and canopy geometry & 3 &
\texttt{none}, \texttt{increase}, \texttt{decrease} &
\texttt{LAI\_max}, \texttt{SLA}, \texttt{n\_plants} \\

2 & Photosynthetic potential & 9 &
\texttt{none}, \texttt{increase}, \texttt{decrease}, \texttt{higher\_sensitivity}, \texttt{lower\_sensitivity} &
\texttt{J\_max\_leaf}, \texttt{Jpot\_activation}, \texttt{Jpot\_deactivation}, \texttt{Jpot\_entropy}, \texttt{alpha}, \texttt{deg\_curv\_elec\_transport}, \texttt{Tcan\_CO2\_comp\_point}, \texttt{CO2\_air\_stomata}, \texttt{net\_ass\_rate} \\

3 & Temperature inhibition & 8 &
\texttt{none}, \texttt{shift\_warm}, \texttt{shift\_cold}, \texttt{widen\_optimum}, \texttt{narrow\_optimum} &
\texttt{k\_sw\_min\_Tcan}, \texttt{s\_min\_Tcan}, \texttt{k\_sw\_max\_Tcan}, \texttt{s\_max\_Tcan}, \texttt{k\_sw\_min\_Tcan24}, \texttt{s\_min\_Tcan24}, \texttt{k\_sw\_max\_Tcan24}, \texttt{s\_max\_Tcan24} \\

4 & Temperature and development & 7 &
\texttt{none}, \texttt{increase}, \texttt{decrease}, \texttt{higher\_sensitivity}, \texttt{lower\_sensitivity} &
\texttt{bias\_g\_Tcan24}, \texttt{slope\_g\_Tcan24}, \texttt{TS\_start}, \texttt{TS\_end}, \texttt{c\_dev1}, \texttt{c\_dev2}, \texttt{r\_fruit\_Set} \\

5 & Biomass growth and maintenance & 12 &
\texttt{none}, \texttt{more\_fruit\_growth}, \texttt{more\_veg\_growth}, \texttt{lower\_resp\_cost}, \texttt{higher\_resp\_cost}, \texttt{higher\_sensitivity}, \texttt{lower\_sensitivity} &
\texttt{G\_max}, \texttt{c\_fruit\_growth}, \texttt{c\_leaf\_growth}, \texttt{c\_stem\_growth}, \texttt{c\_fruit\_maintenance}, \texttt{c\_leaf\_maintenance}, \texttt{c\_stem\_maintenance}, \texttt{Q\_10\_maintenance}, \texttt{rg\_fruit}, \texttt{rg\_leaf}, \texttt{rg\_stem}, \texttt{c\_rgr} \\
\bottomrule
\end{tabularx}
\caption{Grouped model parameters used to construct the discrete adaptation space. Each group is associated with a finite set of canonical perturbations that define coordinated updates over the parameters in that group. The grouping follows the ordered decomposition implemented in the perturbation scheme, from canopy-level traits to biomass growth and maintenance.}
\label{tab:vanthoor_parameter_groups}
\end{table*}

This grouping serves two purposes in the experiments. First, it preserves interpretability by ensuring that each decision can be related to a meaningful component of the mechanistic model. Second, it induces a structured discrete search process in which complete parameterizations are built progressively through a sequence of group-level perturbations. Each state in the search process represents a partially specified simulator parameterization. Starting from a reference configuration, an action applies a discrete perturbation to one parameter group and updates the current model instance accordingly. Because the groups are mutually exclusive, each decision modifies exactly one component of the decomposition and does not overlap with the effects of previous decisions on other groups. A terminal state corresponds to a complete calibration candidate obtained after all required group-level perturbations have been assigned.

Several state-space configurations are considered in order to study both calibration quality and scalability. In the smallest configuration, each parameter group is perturbed once, yielding an enumerable state space of $2625$ terminal states. We refer to this baseline as the 1-cycle setting. This makes it possible to precompute the complete calibration landscape and to compare the learned GFlowNet distribution against the exact reward-defined target distribution. Larger configurations are obtained by allowing repeated perturbation cycles over the same five groups. This preserves the grouped construction principle while increasing the combinatorial size of the terminal-state space, eventually reaching a regime in which exhaustive enumeration is no longer feasible. In the experiments, this larger regime is represented by a 2-cycle setting with \(4.8 \times 10^6\) possible states, and is used to assess the scalability and limitations of our proposed approach under more demanding combinatorial state-spaces.

\subsubsection{Compared Methods and Implementation Details}
\label{sec:experiments-implementation}
The primary method under study is a GFlowNet trained over the grouped discrete perturbation space described above. The forward policy is implemented as a multilayer perceptron with three hidden layers of width 256 and is trained with the trajectory balance objective. Optimization uses a learning rate of \(5\times 10^{-4}\). Because the grouped construction graph is a tree, each terminal state is reached by a unique action sequence, so the backward trajectory probability is fixed to 1 and no learned backward policy is required.

The reward used for training follows the formulation introduced in Section~4.2. For each terminal state, the simulator is evaluated independently on the six observational contexts, the resulting contextual discrepancies are quantile-normalized, and the normalized losses are aggregated into the tail-risk-sensitive adaptation loss \(\mathcal{L}(x)\). In the present experiments, the tail component \(\mathcal{L}_{tail}(x)\) is defined from the two worst-performing contexts, and the robustness weight is fixed to \(\lambda=0.25\). Terminal reward is then obtained through the Boltzmann construction \(R(x)=\exp(-\beta \mathcal{L}(x))\), so that the concentration of probability mass around low-loss regions is controlled by the parameter \(\beta\).

Two experimental factors are varied systematically. The first is the grouped perturbation magnitude, controlled through the step fraction, for which we consider \(sf=0.15\) and \(sf=0.3\), corresponding respectively to finer and coarser grouped moves in parameter space. The second is the reward concentration parameter \(\beta\). In the 1-cycle setting, we evaluate \(\beta \in \{2,4,8\}\). In the 2-cycle setting, we retain the best 1-cycle value, \(\beta=4\), since a full ablation becomes prohibitively expensive once simulator evaluations are no longer cached. For repeated perturbation cycles, the effective perturbation magnitude is annealed across cycles by halving the update size at each pass over the parameter groups. This preserves the grouped construction principle while allowing later cycles to refine earlier decisions more conservatively.

Training budgets are adjusted to reflect the computational difference between the enumerable and non-enumerable settings. In the 1-cycle setting, the full reward landscape can be precomputed and cached, so training scales well and could in principle be extended further; in practice, models are trained for 1000 steps with a batch size of 16. In the 2-cycle setting, simulator evaluations become the main bottleneck, so training is limited to 2000 steps as a compromise between convergence and wall-clock cost. This larger setting is therefore used not only to test scalability, but also to expose the practical limits of the current adaptation mechanism under realistic simulation-evaluation constraints.

Two baseline search methods are used for comparison. The first is random search over the same grouped discrete perturbation structure as the GFlowNet. The second is Optuna's Tree-structured Parzen Estimator~\cite{watanabe2023tree, akiba2019optuna} (TPE) adapted to the same grouped discrete search space. TPE was chosen as a strong adaptive baseline because it performs well in structured discrete optimization and provides a competitive alternative to unguided exploration, as it was designed for structured discrete spaces. For retrieval comparisons, methods are evaluated under budgets that reflect the computational regime being studied. In the enumerable 1-cycle setting, comparisons are made under matched sampling budgets, since accounting for GFlowNet training cost would allow adaptive baselines to cover a large fraction of the finite search space directly. In the 2-cycle setting, where simulator evaluation becomes the dominant bottleneck, methods are compared under matched simulator-evaluation budgets to assess scalability.

%% file: sections/experiments/results.tex
\subsection{Results}
\label{sec:results}
This subsection reports the empirical results with respect to the three research
questions introduced in Section~\ref{sec:eval}. We begin with mode discovery in the enumerable
1-cycle regime, where the exact reward landscape makes it possible to compare the
learned GFlowNet distribution against the target landscape and its dominant basins.
We then examine retrieval performance under fixed budgets through ranked
distributional comparisons, top-$k$ recovery, and best-so-far behavior. Finally, we
turn to the larger 2-cycle regime, where exhaustive analysis is no longer feasible and
evaluation focuses instead on retrieval quality, diversity among strong retrieved
candidates, and qualitative agreement with the observed trajectories. Together, these
results assess whether the proposed generative adaptation mechanism can recover
high-quality regions of the adaptation space, retrieve strong calibration hypotheses,
and remain useful as simulator-evaluation costs increase.

\subsubsection{Mode discovery}
\label{sec:results-rq1}
\textbf{RQ1} examines whether the proposed GFlowNet recovers distinct high-quality regions of the adaptation landscape, thereby enabling the generation of diverse and plausible simulator parameterizations under sparse observations. This question is studied in the enumerable 1-cycle regime, where the exact reward landscape is available and can be used as ground truth. We focus here on the representative setting \(\beta=4\), and compare two step fractions, \(sf=0.15\) and \(sf=0.3\), in order to show how the structure of the adaptation landscape changes with the granularity of the grouped perturbations.

\begin{figure}[H]
    \centering

    \begin{minipage}[t]{0.32\textwidth}
        \centering
        \includegraphics[width=\linewidth]{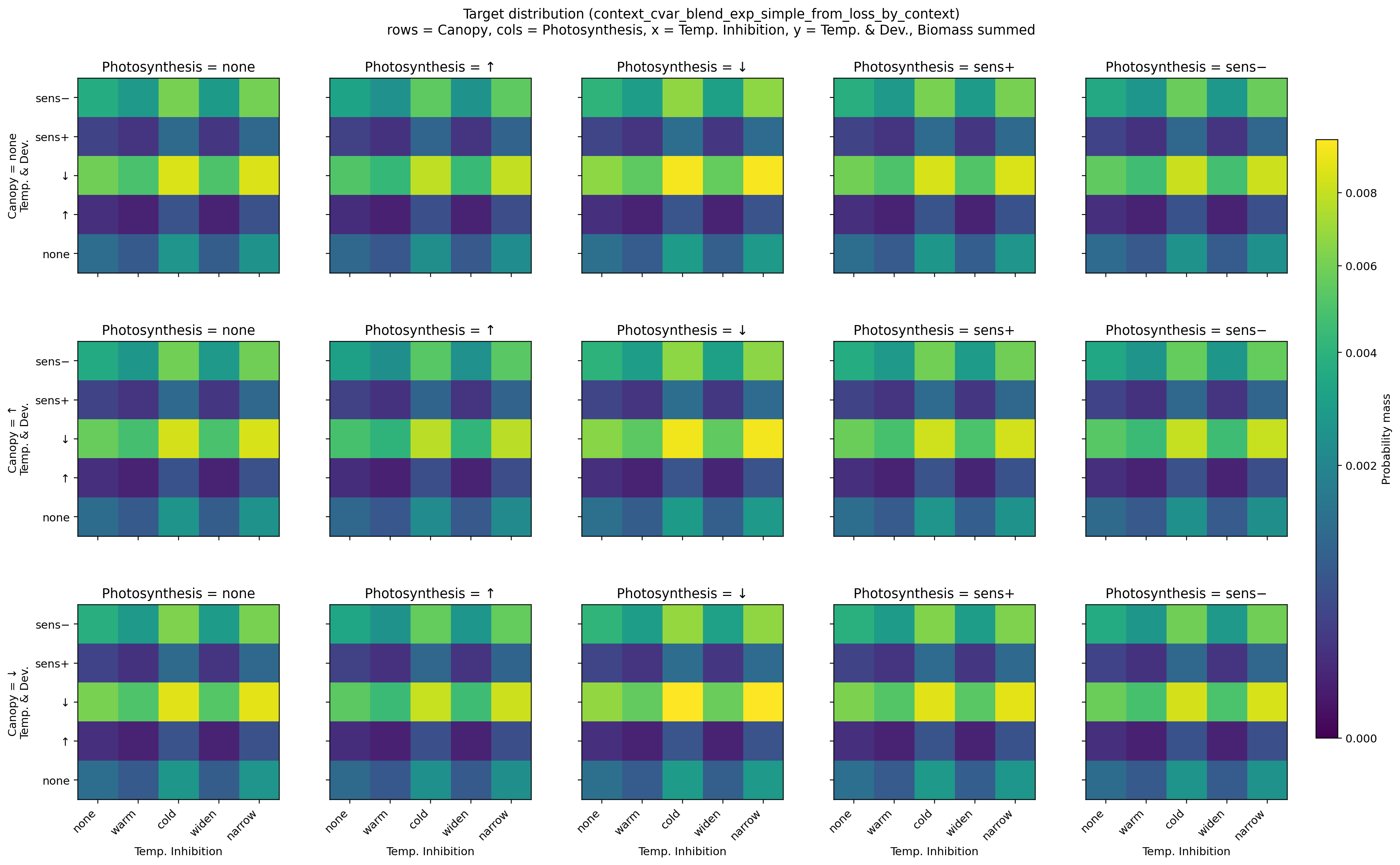}
        
        \small (a) Exact target distribution, \(sf=0.15\)
    \end{minipage}
    \hfill
    \begin{minipage}[t]{0.32\textwidth}
        \centering
        \includegraphics[width=\linewidth]{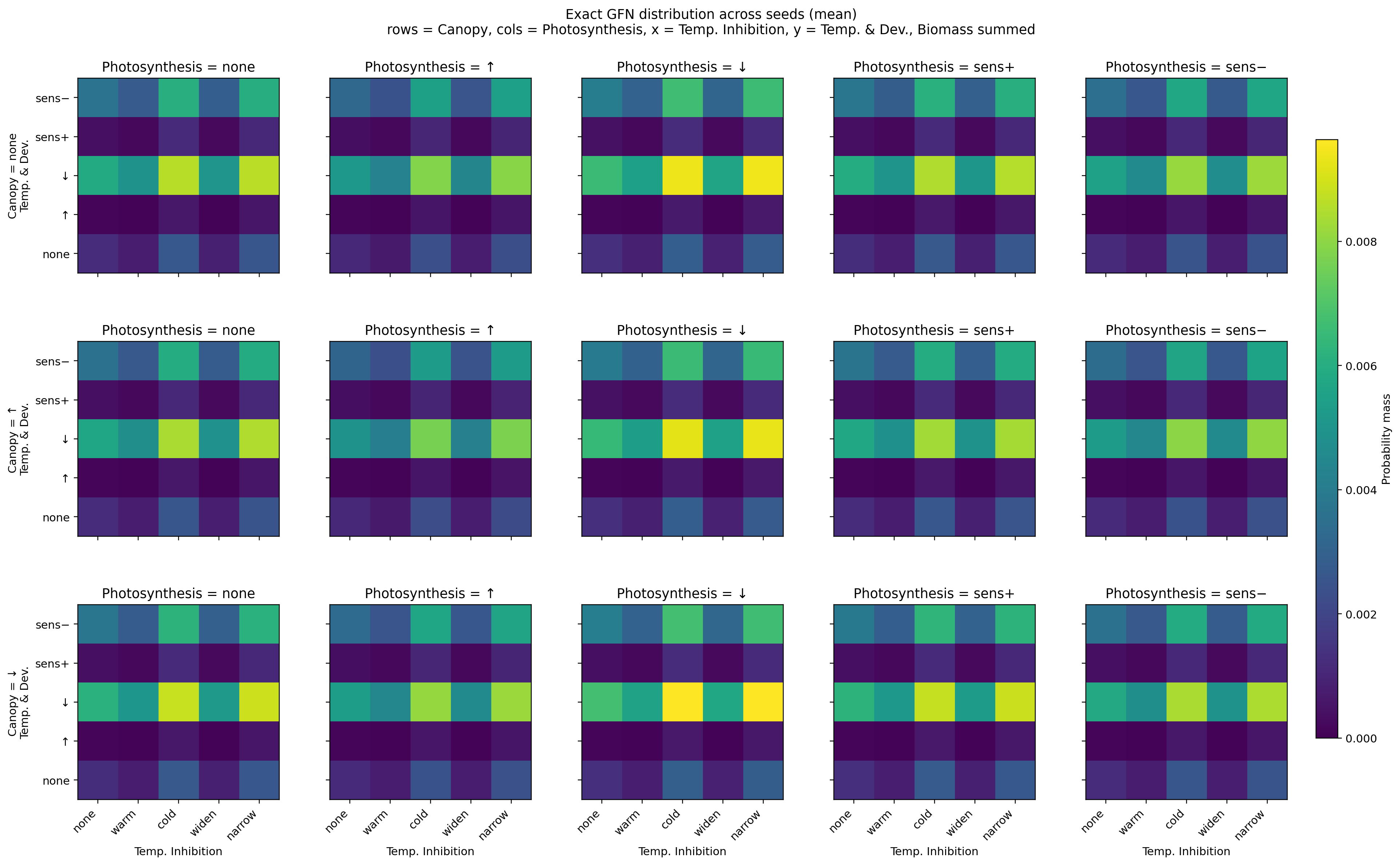}
        
        \small (b) Mean GFlowNet distribution, \(sf=0.15\)
    \end{minipage}
    \hfill
    \begin{minipage}[t]{0.32\textwidth}
        \centering
        \includegraphics[width=\linewidth]{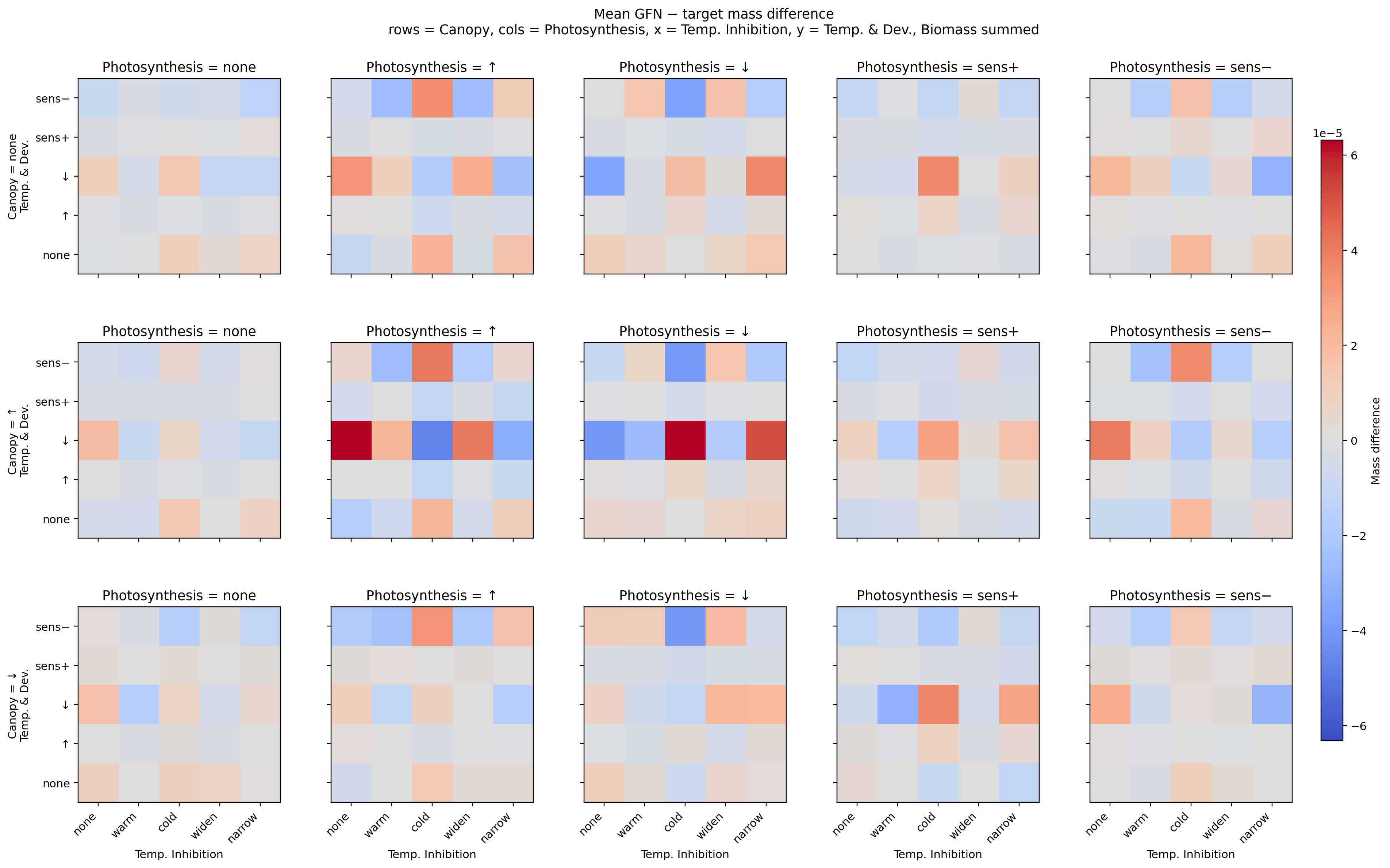}
        
        \small (c) Mass difference, \(sf=0.15\)
    \end{minipage}

    \vspace{0.6em}

    \begin{minipage}[t]{0.32\textwidth}
        \centering
        \includegraphics[width=\linewidth]{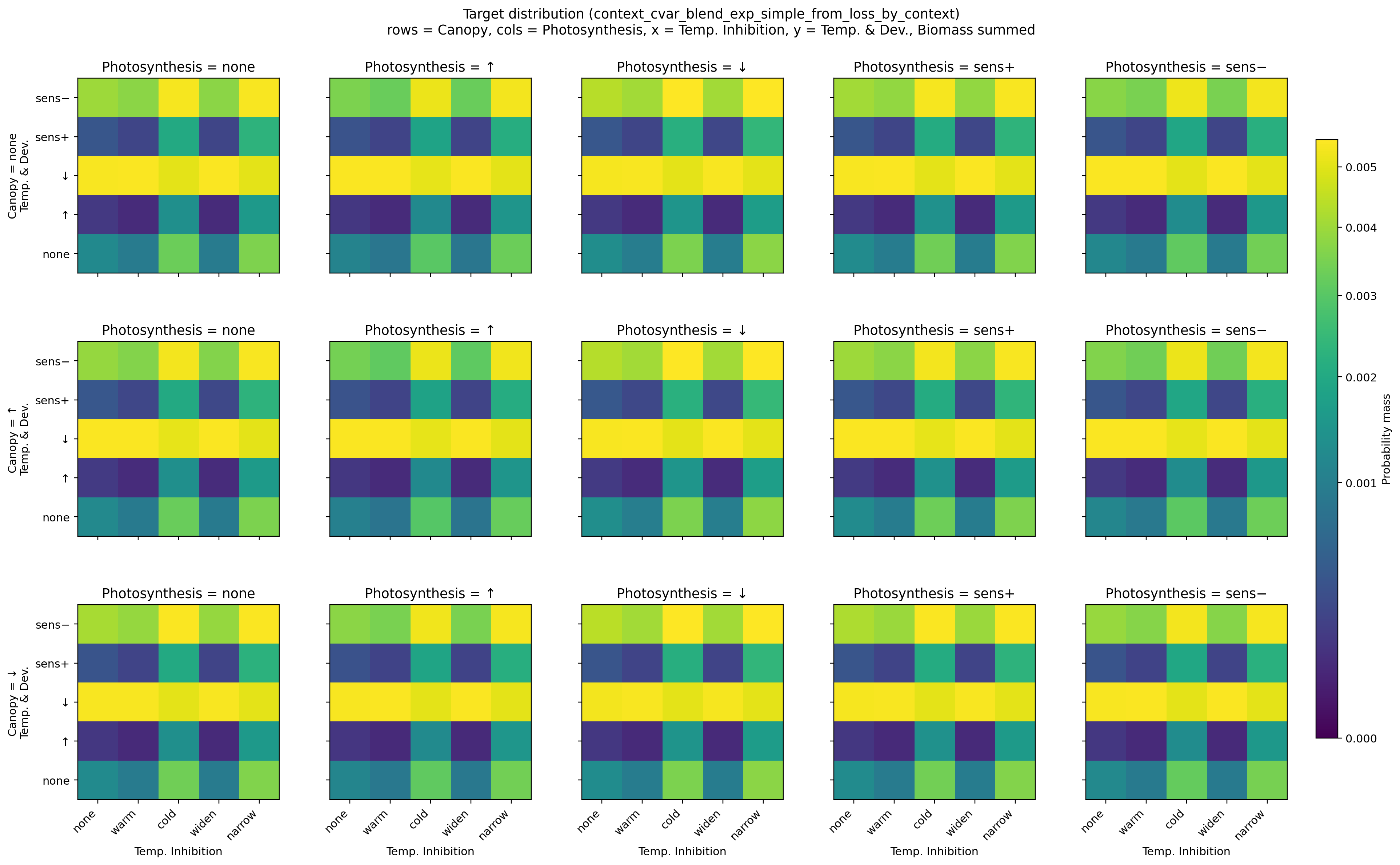}
        
        \small (d) Exact target distribution, \(sf=0.3\)
    \end{minipage}
    \hfill
    \begin{minipage}[t]{0.32\textwidth}
        \centering
        \includegraphics[width=\linewidth]{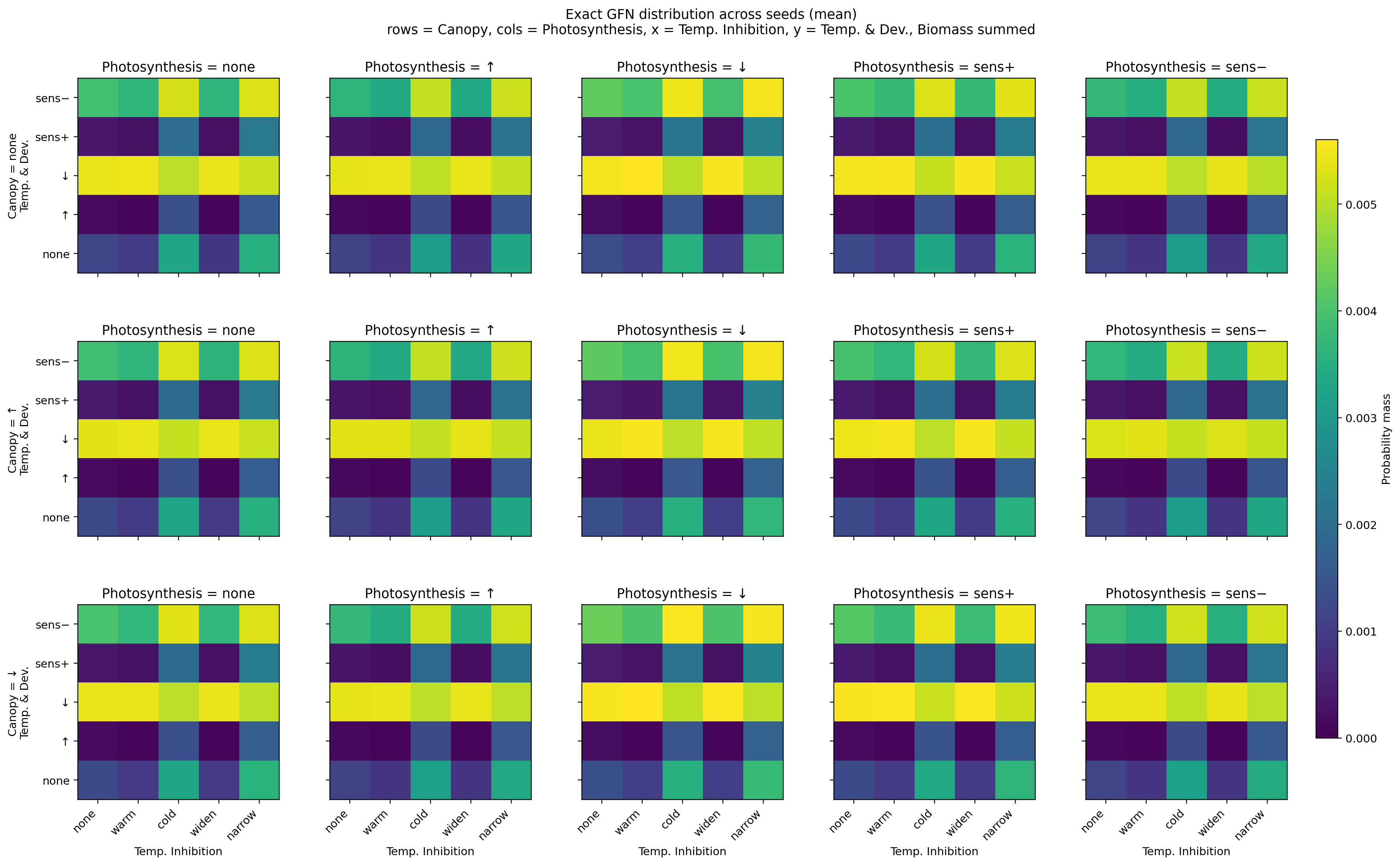}
        
        \small (e) Mean GFlowNet distribution, \(sf=0.3\)
    \end{minipage}
    \hfill
    \begin{minipage}[t]{0.32\textwidth}
        \centering
        \includegraphics[width=\linewidth]{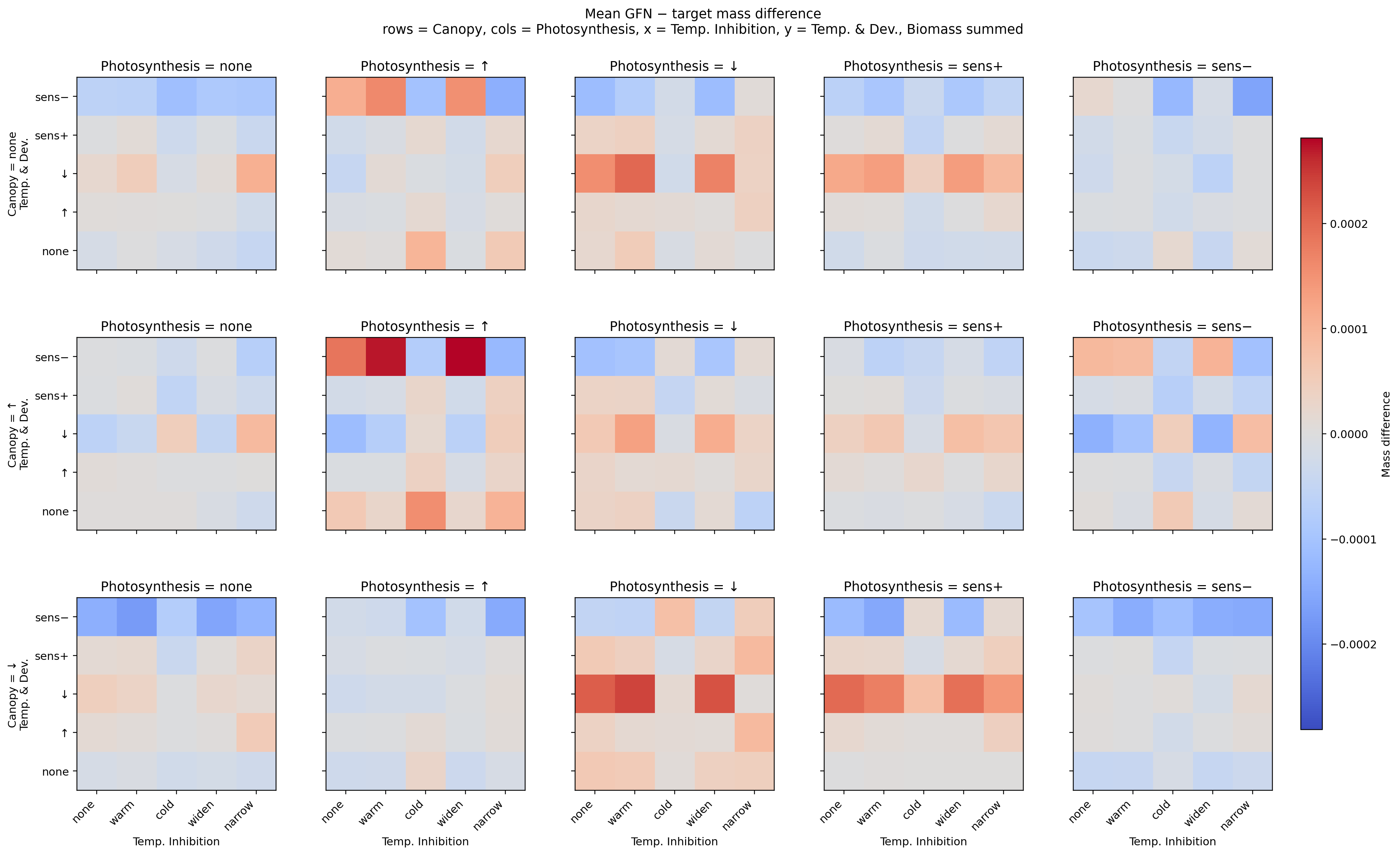}
        
        \small (f) Mass difference, \(sf=0.3\)
    \end{minipage}

    \caption{Mode-discovery analysis in the enumerable 1-cycle setting for \(\beta=4\) under two step fractions. Panels (a)--(c) correspond to the finer perturbation regime \(sf=0.15\), while panels (d)--(f) correspond to the coarser regime \(sf=0.3\). In each row, the left panel shows the exact target distribution, the middle panel shows the mean GFlowNet terminal distribution across seeds, and the right panel shows the difference between the two.}
    \label{fig:rq1_landscape_compare}
\end{figure}

\Fig{fig:rq1_landscape_compare} compares the exact target distribution, the mean GFlowNet terminal distribution across seeds, and the corresponding mass difference for the two step fractions. In both settings, the learned policy recovers the overall structure of the exact target distribution and places mass on the same dominant high-reward regions. We observe that the GFlowNet tends to concentrate slightly more mass on the highest-reward areas of the landscape, while assigning less mass to lower-reward states. This effect is limited in the \(sf=0.15\) setting, where the target landscape is largely dominated by a single basin, and remains localized in the \(sf=0.3\) setting, where the landscape becomes genuinely multimodal. Overall, the learned distribution closely tracks the exact target while exhibiting a mild preference for its most rewarding regions.

\begin{figure}[H]
    \centering
    \begin{minipage}[t]{0.48\textwidth}
        \centering
        \includegraphics[width=\linewidth]{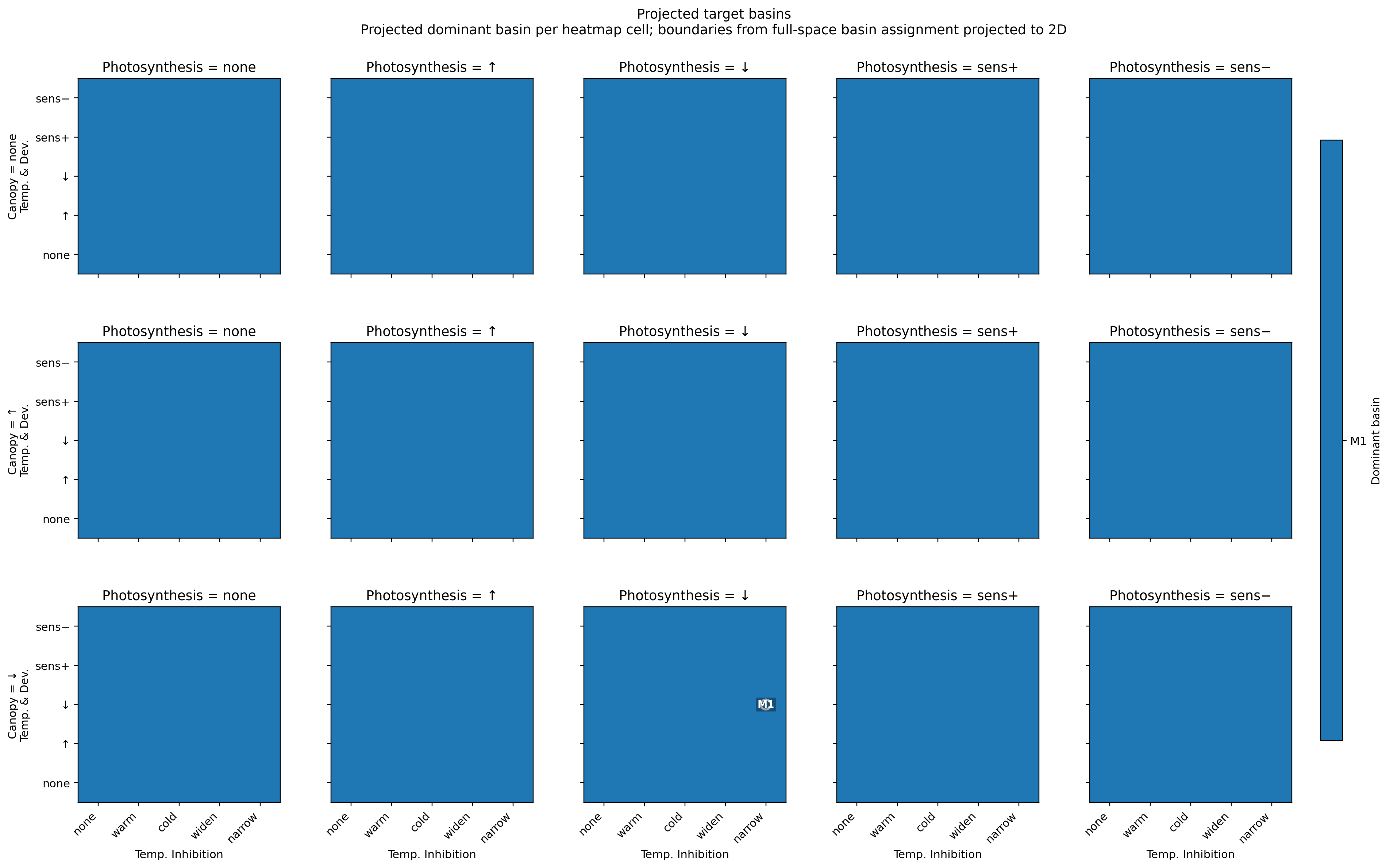}
        
        \small (a) Basin map for \(sf=0.15,\ \beta=4\)
    \end{minipage}
    \hfill
    \begin{minipage}[t]{0.48\textwidth}
        \centering
        \includegraphics[width=\linewidth]{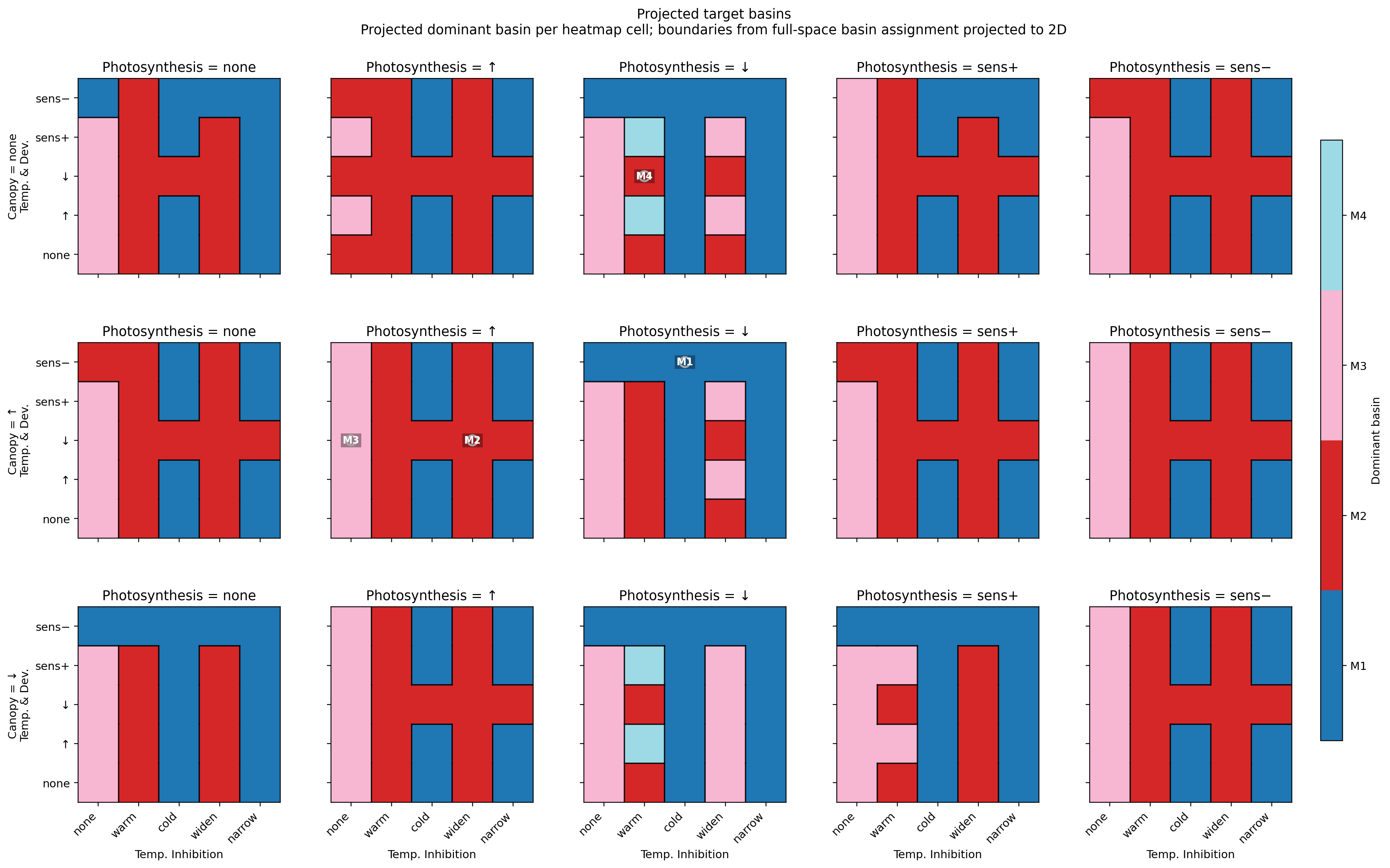}
        
        \small (b) Basin map for \(sf=0.3,\ \beta=4\)
    \end{minipage}
    \caption{Comparison of basin structure in the enumerable 1-cycle setting for \(\beta=4\) under two step fractions. Basins are obtained from the exact enumerable landscape by assigning each terminal state to the local mode reached by ascent on the full 1-cycle neighborhood graph. The projected maps show, for each displayed cell, the dominant basin contributing the largest share of mass in that region.}
    \label{fig:rq1_basin_maps_compare}
\end{figure}

Modes are showcased more explicitly in \Fig{fig:rq1_basin_maps_compare}. Here, basins are not defined directly in the projected heatmaps. Instead, they are obtained from the exact enumerable 1-cycle landscape by representing terminal states as a neighborhood graph and assigning each state to the local mode reached by repeated ascent toward higher target probability. Basin mass is then obtained by summing the probability mass of all member states assigned to that mode. The basin comparison shows that the finer step fraction yields a landscape dominated by a single basin, whereas the coarser step fraction reveals four dominant basins. This difference helps explain why the \(sf=0.3\) setting is the most informative one for evaluating mode discovery.

\begin{figure}[H]
    \centering

    \begin{minipage}[t]{0.32\textwidth}
        \centering
        \includegraphics[width=\linewidth]{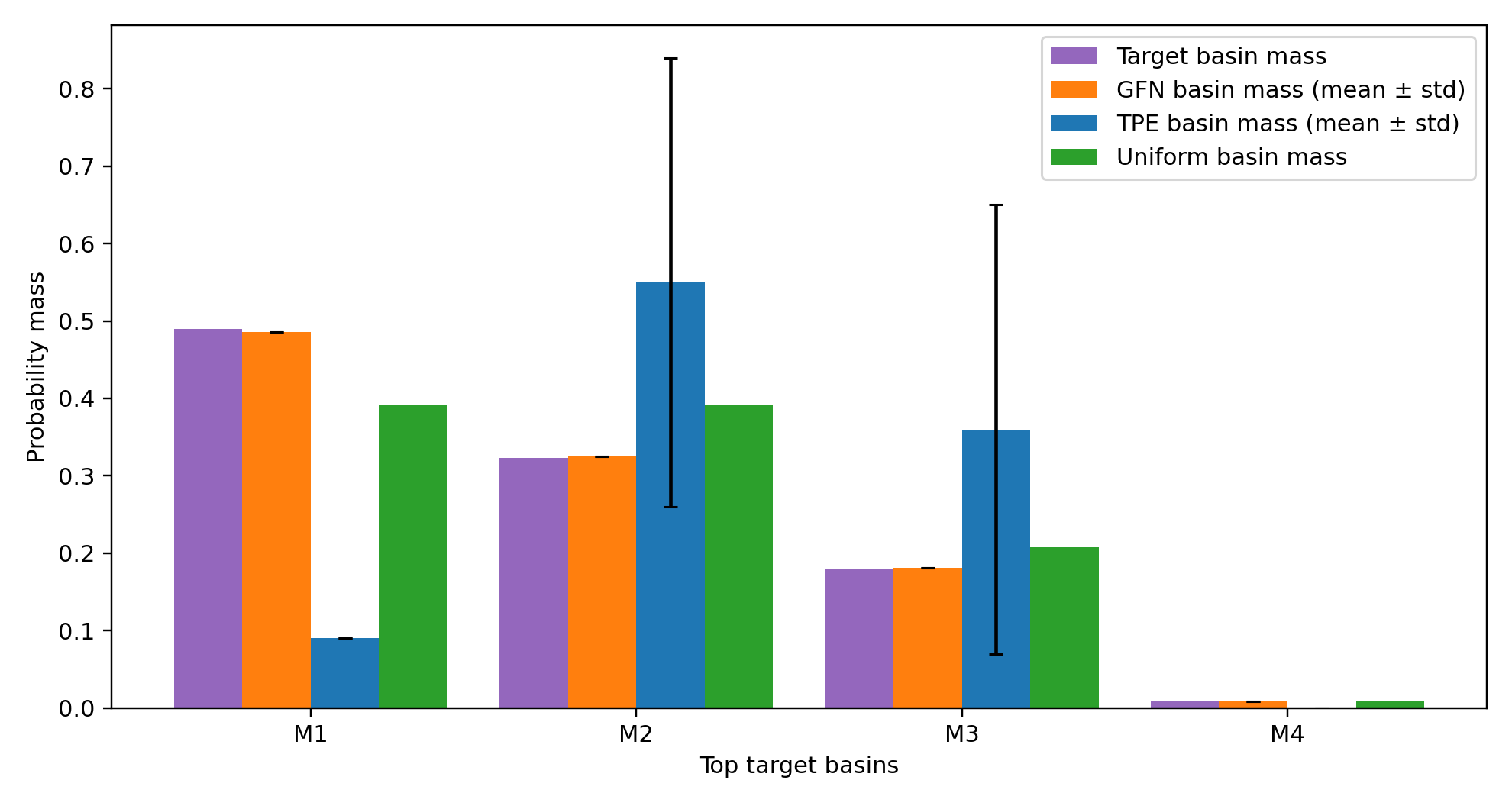}
        
        \small (a) Basin masses for \(sf=0.3,\ \beta=2\)
    \end{minipage}
    \hfill
    \begin{minipage}[t]{0.32\textwidth}
        \centering
        \includegraphics[width=\linewidth]{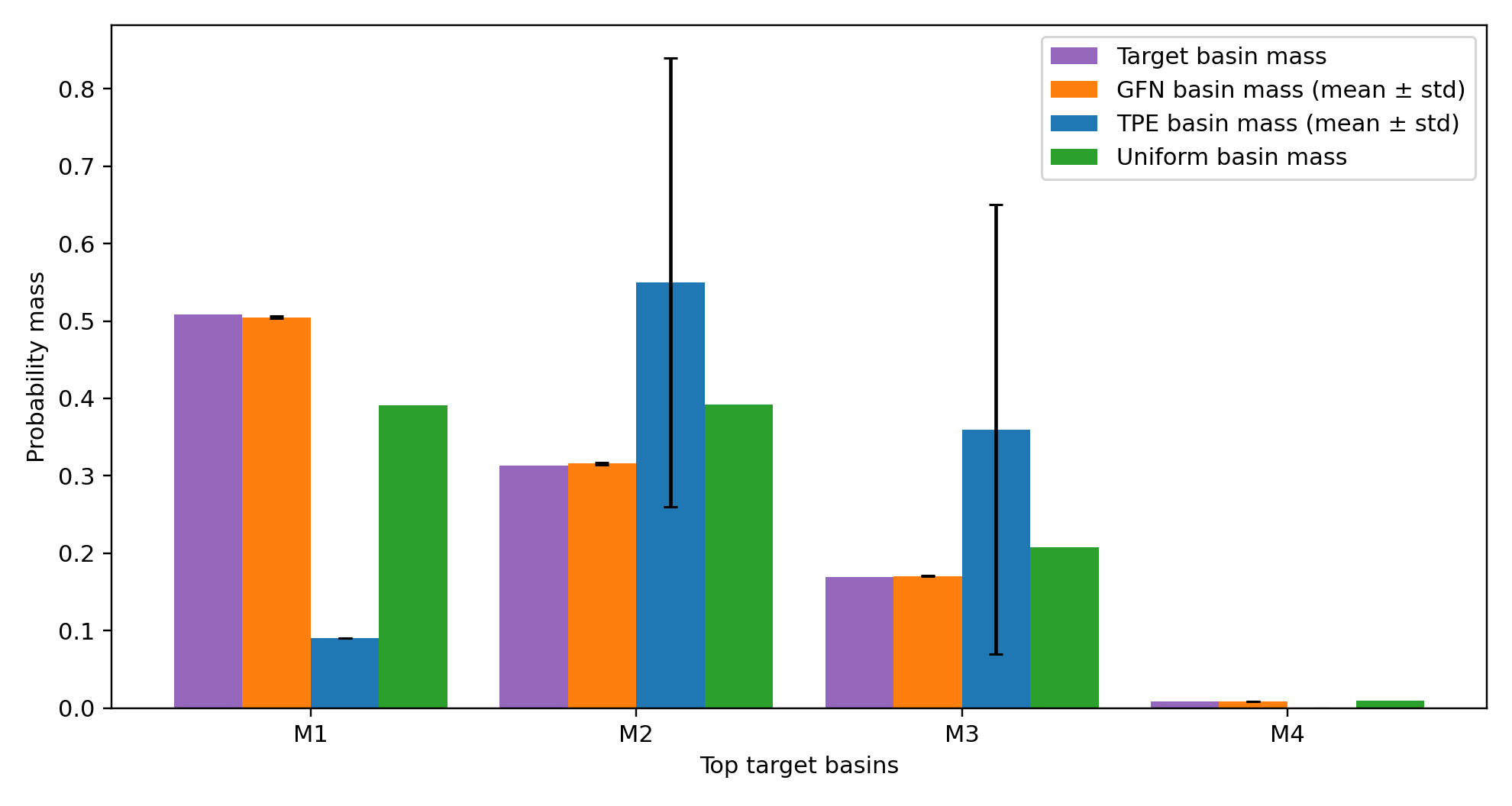}
        
        \small (b) Basin masses for \(sf=0.3,\ \beta=4\)
    \end{minipage}
    \hfill
    \begin{minipage}[t]{0.32\textwidth}
        \centering
        \includegraphics[width=\linewidth]{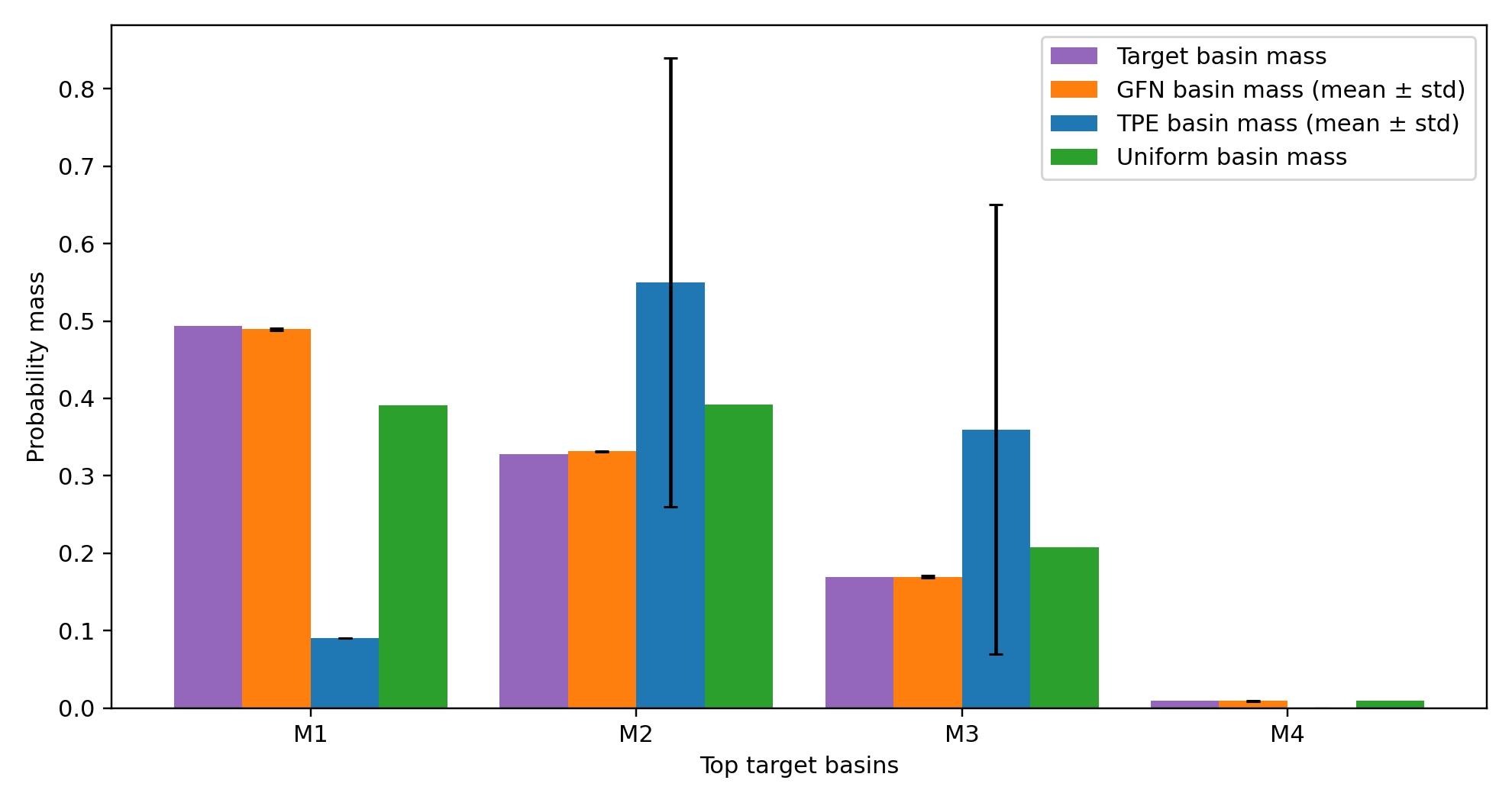}
        
        \small (c) Basin masses for \(sf=0.3,\ \beta=8\)
    \end{minipage}

    \caption{Comparison of probability mass assigned to the dominant target basins in the enumerable 1-cycle setting for \(sf=0.3\) under three reward sharpness settings. From left to right, the panels show \(\beta=2\), \(\beta=4\), and \(\beta=8\).}
    \label{fig:rq1_basin_masses_compare}
\end{figure}

The basin-level comparison in \Fig{fig:rq1_basin_masses_compare} shows that the proposed approach consistently recovers the dominant target modes across all three reward sharpness settings. In other words, mode discovery is not restricted to a particular choice of \(\beta\). The overall concentration of probability mass across the principal basins remains broadly similar from \(\beta=2\) to \(\beta=8\), which suggests that changing the reward sharpness has only a limited effect on the basin-level allocation itself. What remains stable across these settings is that the GFlowNet places mass on the same dominant basins as the exact target distribution, rather than collapsing onto a single optimum or dispersing mass arbitrarily. This provides further evidence that the learned policy captures the principal mode structure of the adaptation landscape.

Taken together, these results show that the proposed approach preserves multiple plausible high-quality parameterizations rather than collapsing onto a single solution. This is especially clear when the grouped perturbation space induces a genuinely multimodal target distribution, as in the \(sf=0.3\) setting, where the GFlowNet recovers the dominant basins of the exact landscape. In that sense, these results provide evidence for \rtwo{}: the learned policy does not merely identify one strong calibration, but maintains several plausible regions of the parameter space from which diverse simulator configurations can be sampled.

\subsubsection{Retrieval performance}
\label{sec:results-rq2}
\textbf{RQ2} investigates whether the learned GFlowNet policy is useful not only as a generative model of the adaptation landscape, but also as a practical retrieval mechanism for high-quality simulator parameterizations. This question is again studied in the enumerable 1-cycle regime, where the exact target distribution is available and can be used as a reference. We evaluate retrieval from two complementary perspectives. First, we compare the ranked probability profiles induced by the learned and exact distributions in order to assess whether the GFlowNet preserves the global allocation of mass across terminal states. Second, we measure retrieval quality directly through top-\(k\) recovery, best-so-far performance, and qualitative overlays against the observed dry-mass measurements.

\begin{figure}[H]
    \centering

    \begin{minipage}[t]{0.32\textwidth}
        \centering
        \includegraphics[width=\linewidth]{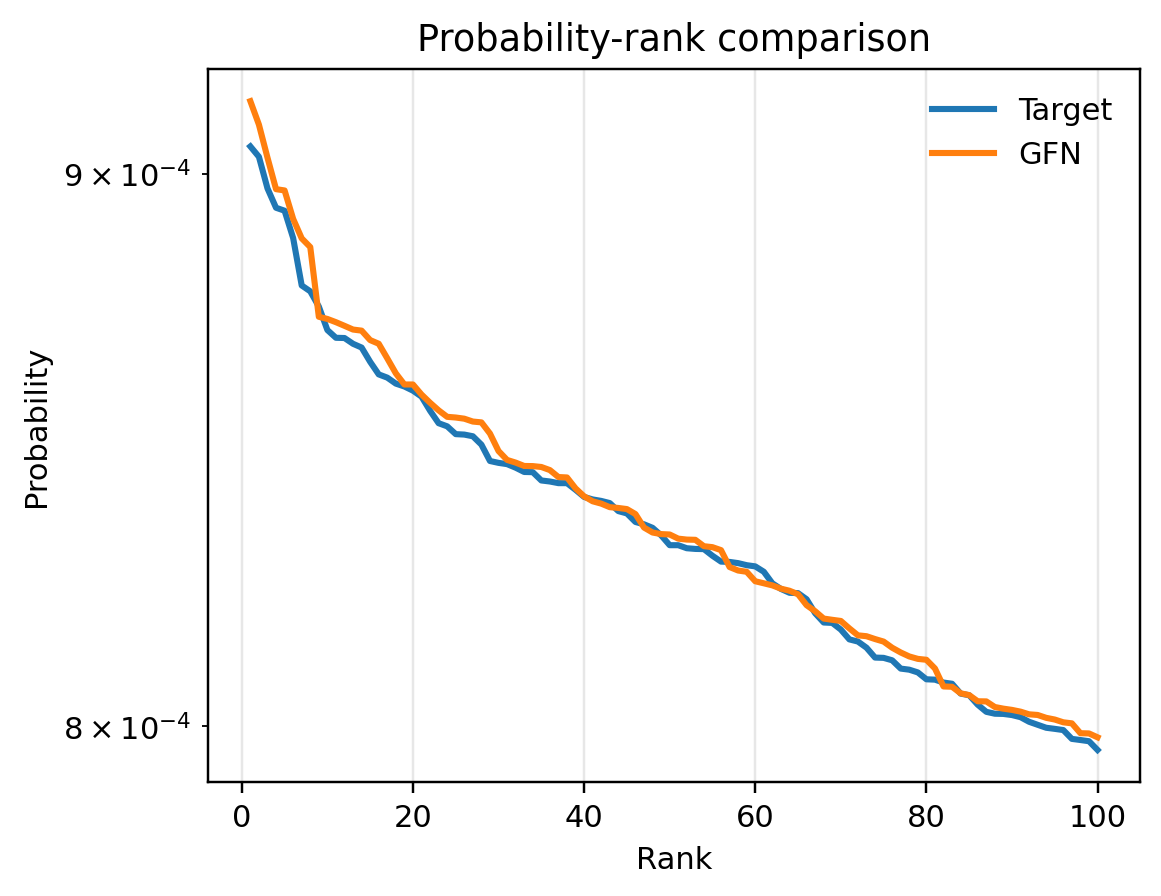}
        
        \small (a) \(sf=0.15,\ \beta=2\)
    \end{minipage}
    \hfill
    \begin{minipage}[t]{0.32\textwidth}
        \centering
        \includegraphics[width=\linewidth]{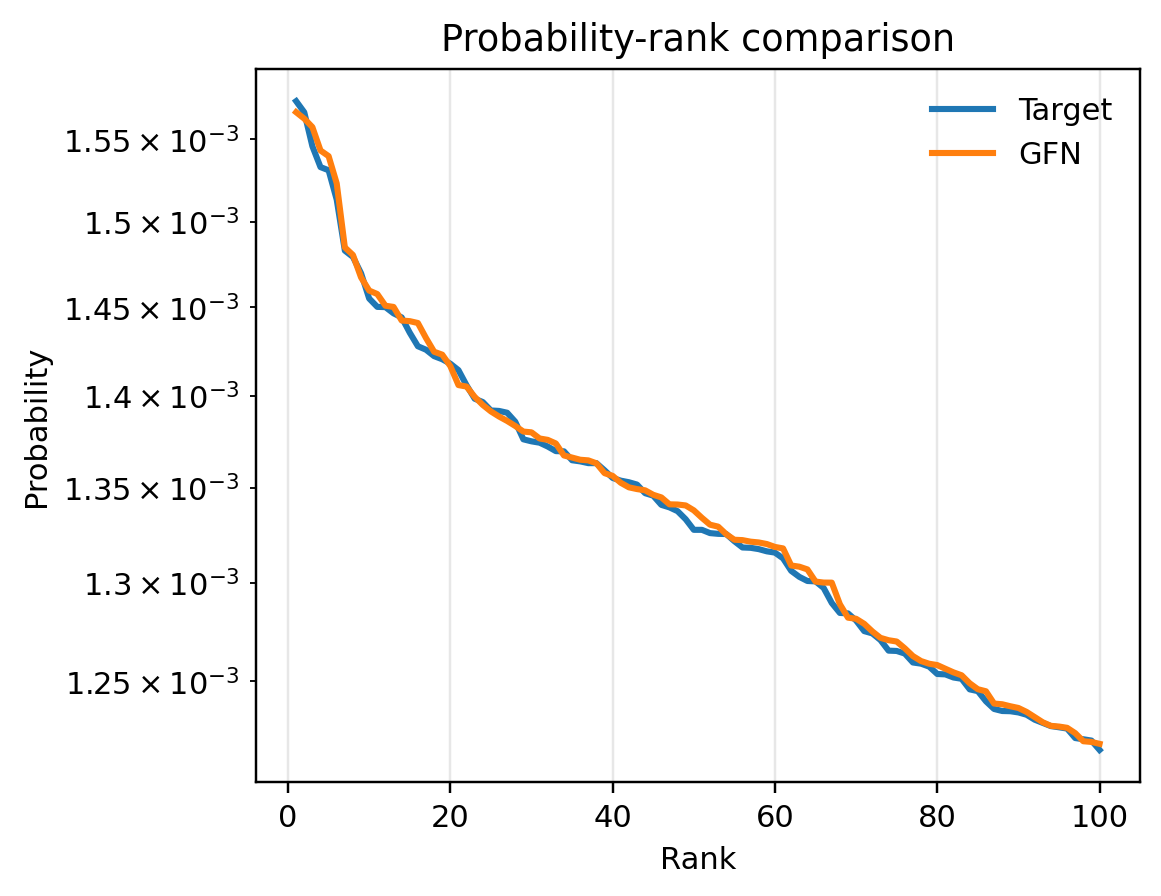}
        
        \small (b) \(sf=0.15,\ \beta=4\)
    \end{minipage}
    \hfill
    \begin{minipage}[t]{0.32\textwidth}
        \centering
        \includegraphics[width=\linewidth]{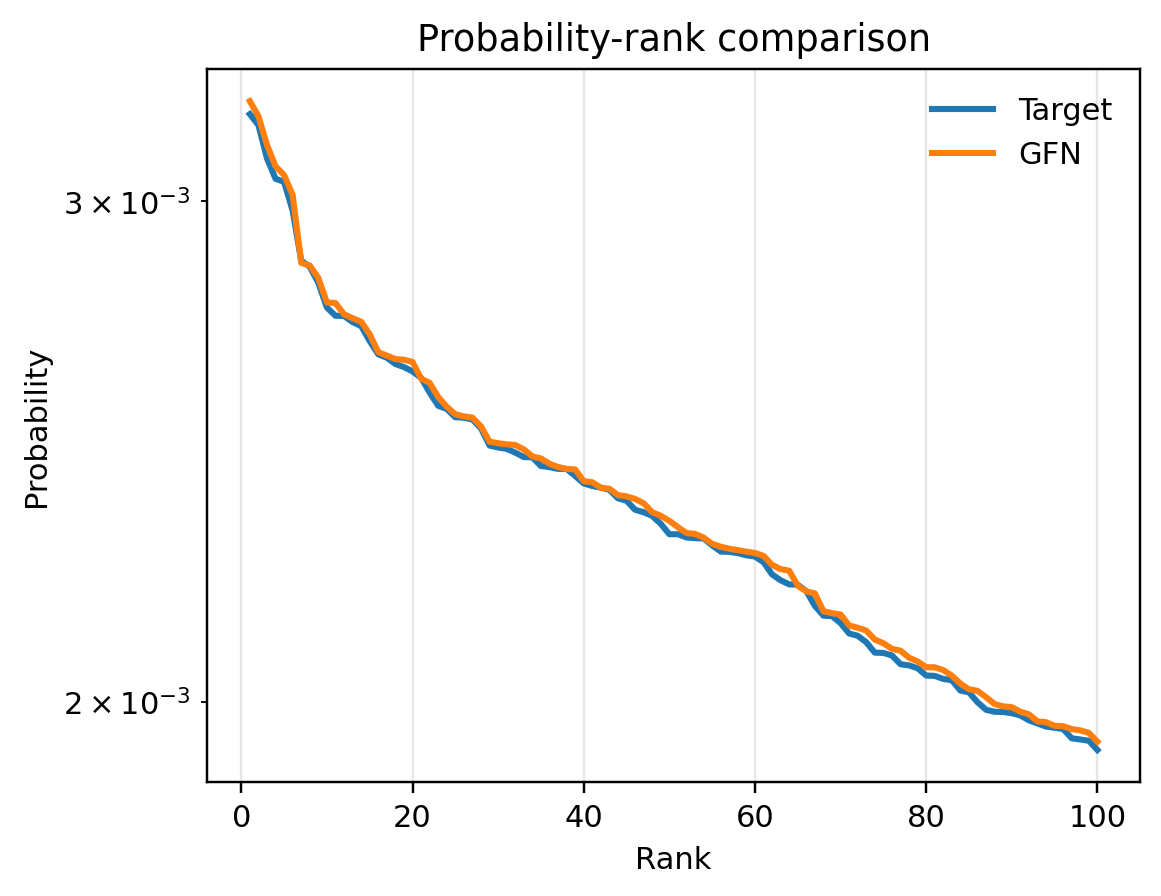}
        
        \small (c) \(sf=0.15,\ \beta=8\)
    \end{minipage}

    \vspace{0.6em}

    \begin{minipage}[t]{0.32\textwidth}
        \centering
        \includegraphics[width=\linewidth]{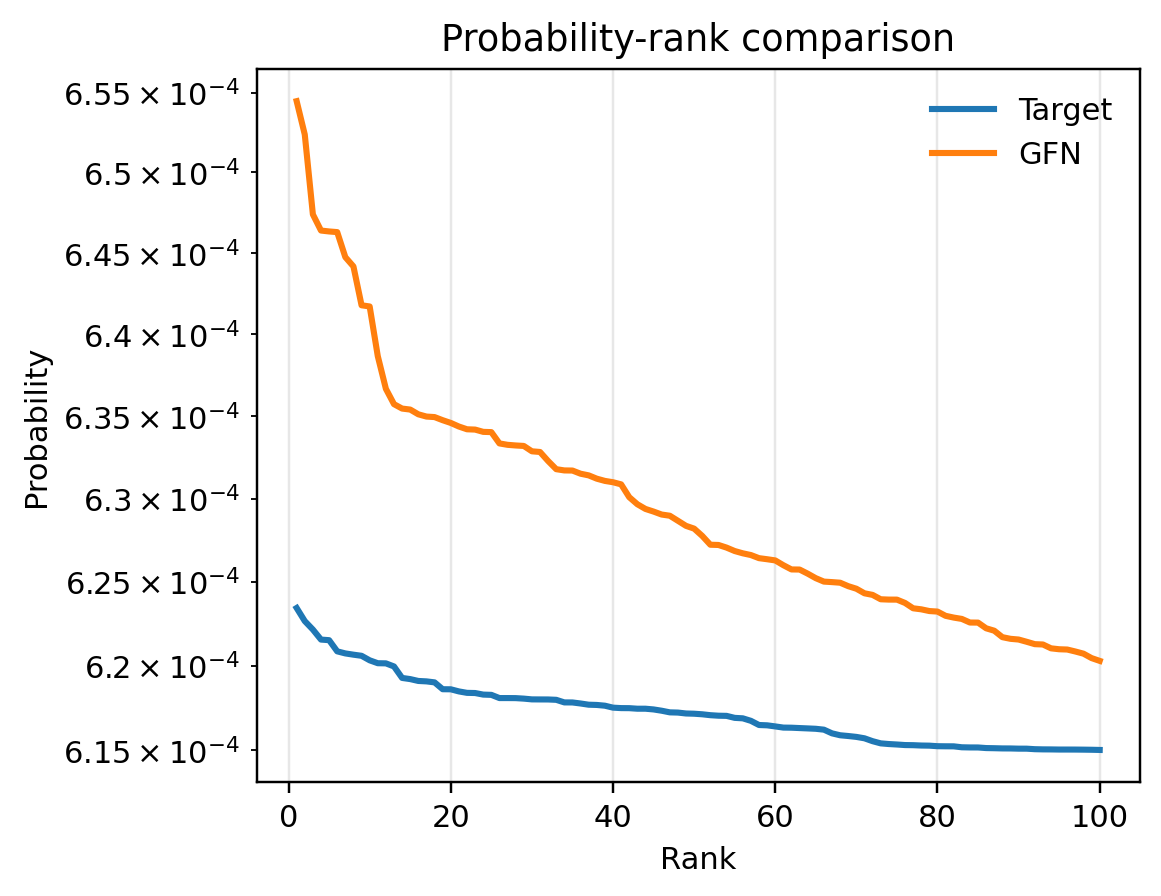}
        
        \small (d) \(sf=0.3,\ \beta=2\)
    \end{minipage}
    \hfill
    \begin{minipage}[t]{0.32\textwidth}
        \centering
        \includegraphics[width=\linewidth]{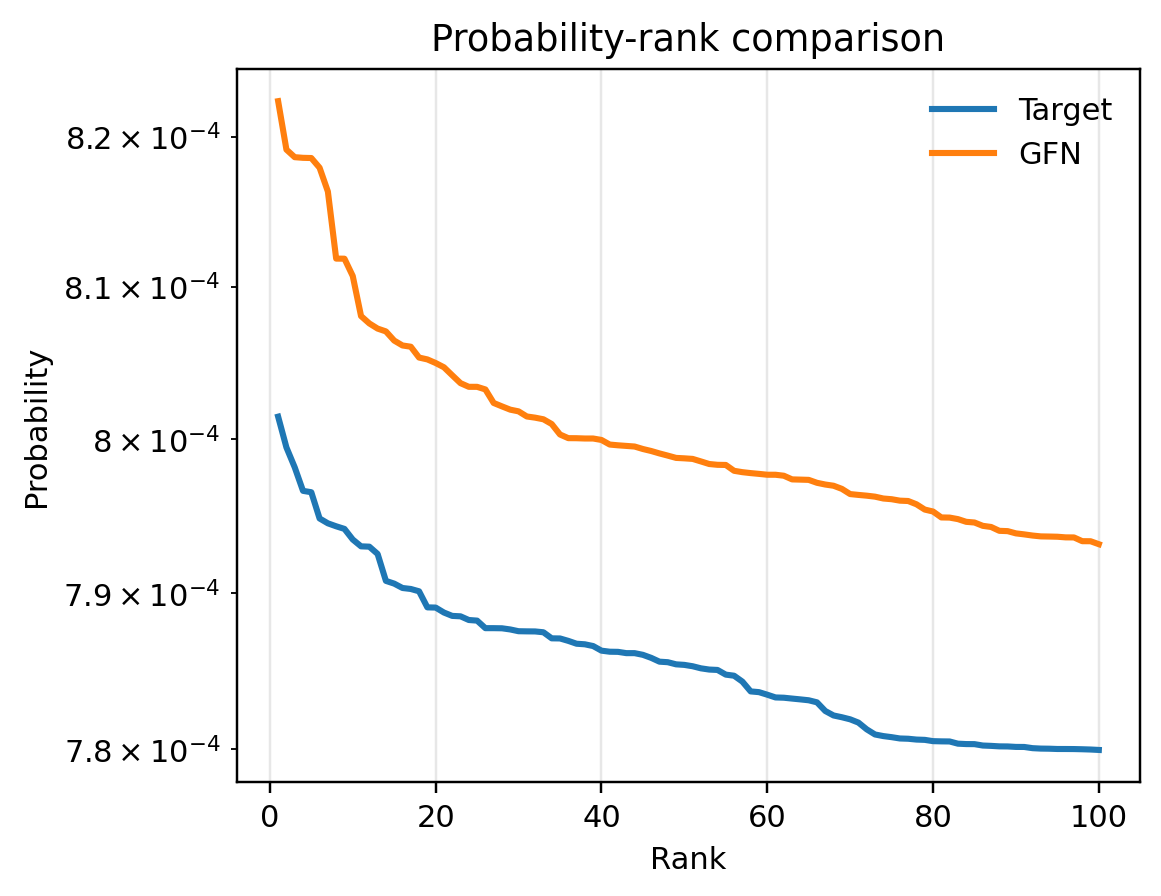}
        
        \small (e) \(sf=0.3,\ \beta=4\)
    \end{minipage}
    \hfill
    \begin{minipage}[t]{0.32\textwidth}
        \centering
        \includegraphics[width=\linewidth]{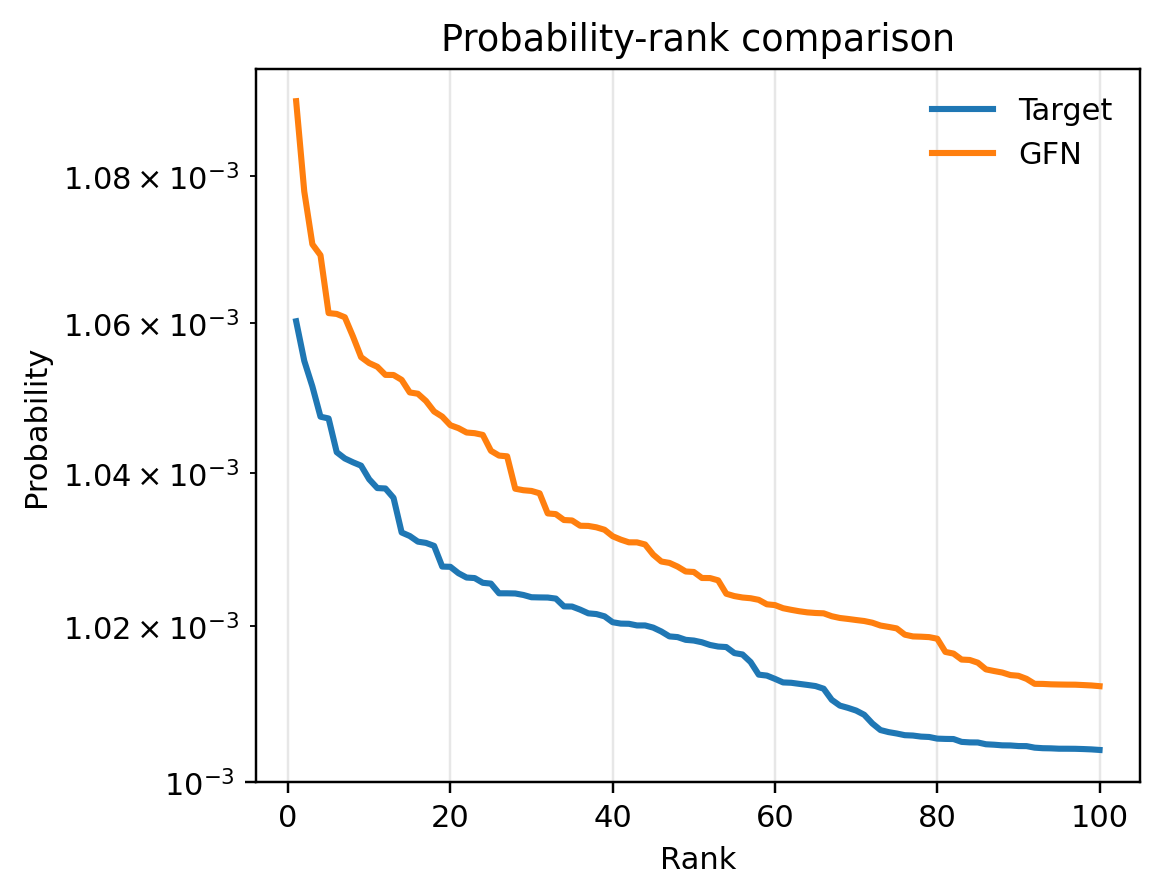}
        
        \small (f) \(sf=0.3,\ \beta=8\)
    \end{minipage}

    \caption{Probability-rank comparison between the exact target distribution and the learned GFlowNet distribution in the enumerable 1-cycle setting under two step fractions and three reward sharpness settings. In each panel, states are ordered by decreasing probability and the ranked profiles of the two distributions are compared directly.}
    \label{fig:rq2_probability_rank_compare}
\end{figure}

\Fig{fig:rq2_probability_rank_compare} shows that the learned distribution remains closely aligned with the ranked probability structure of the exact target distribution across all six settings. This indicates that the GFlowNet preserves not only the dominant modes identified in RQ1, but also the broader allocation of mass across high-quality terminal states. For \(sf=0.15\), this agreement is strong for all three values of \(\beta\), and the learned distribution closely matches the exact target distribution. For \(sf=0.3\), the match becomes less exact, which suggests that the learned normalization constant \(Z\) shifts the overall allocation of probability mass. However, even in that case the proportionality to the underlying reward landscape is preserved. The ranked comparisons show that the learned policy remains globally consistent with the exact target distribution, with the main deviations appearing at the level of normalization rather than in the ordering of high-quality states. As a result, the highest-scoring candidates receive more probability mass under the learned distribution than they would under the exact target landscape, and are therefore sampled more often.

\begin{figure}[H]
    \centering

    \begin{minipage}[t]{0.32\textwidth}
        \centering
        \includegraphics[width=\linewidth]{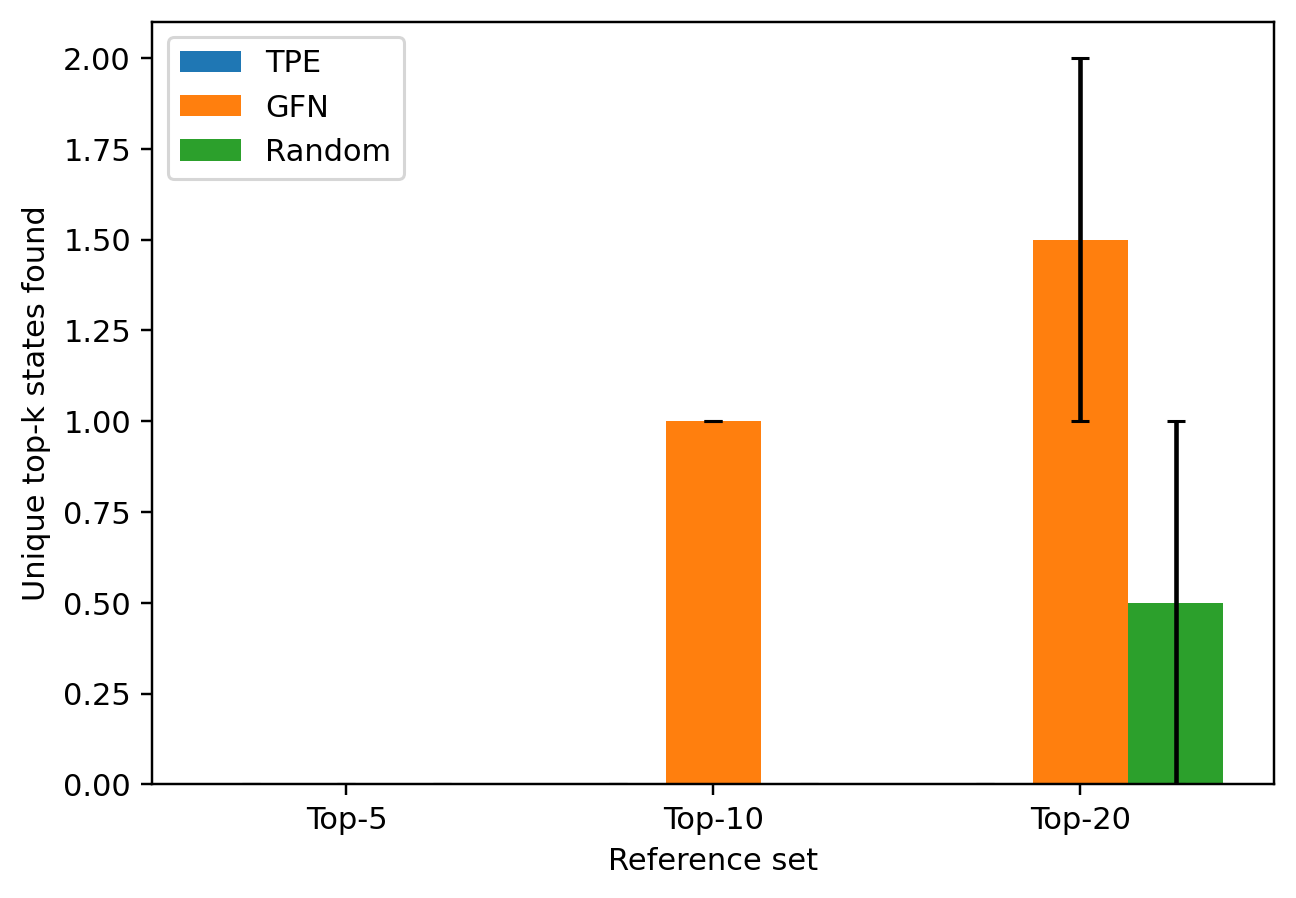}
        
        \small (a) \(sf=0.3,\ \beta=2\)
    \end{minipage}
    \hfill
    \begin{minipage}[t]{0.32\textwidth}
        \centering
        \includegraphics[width=\linewidth]{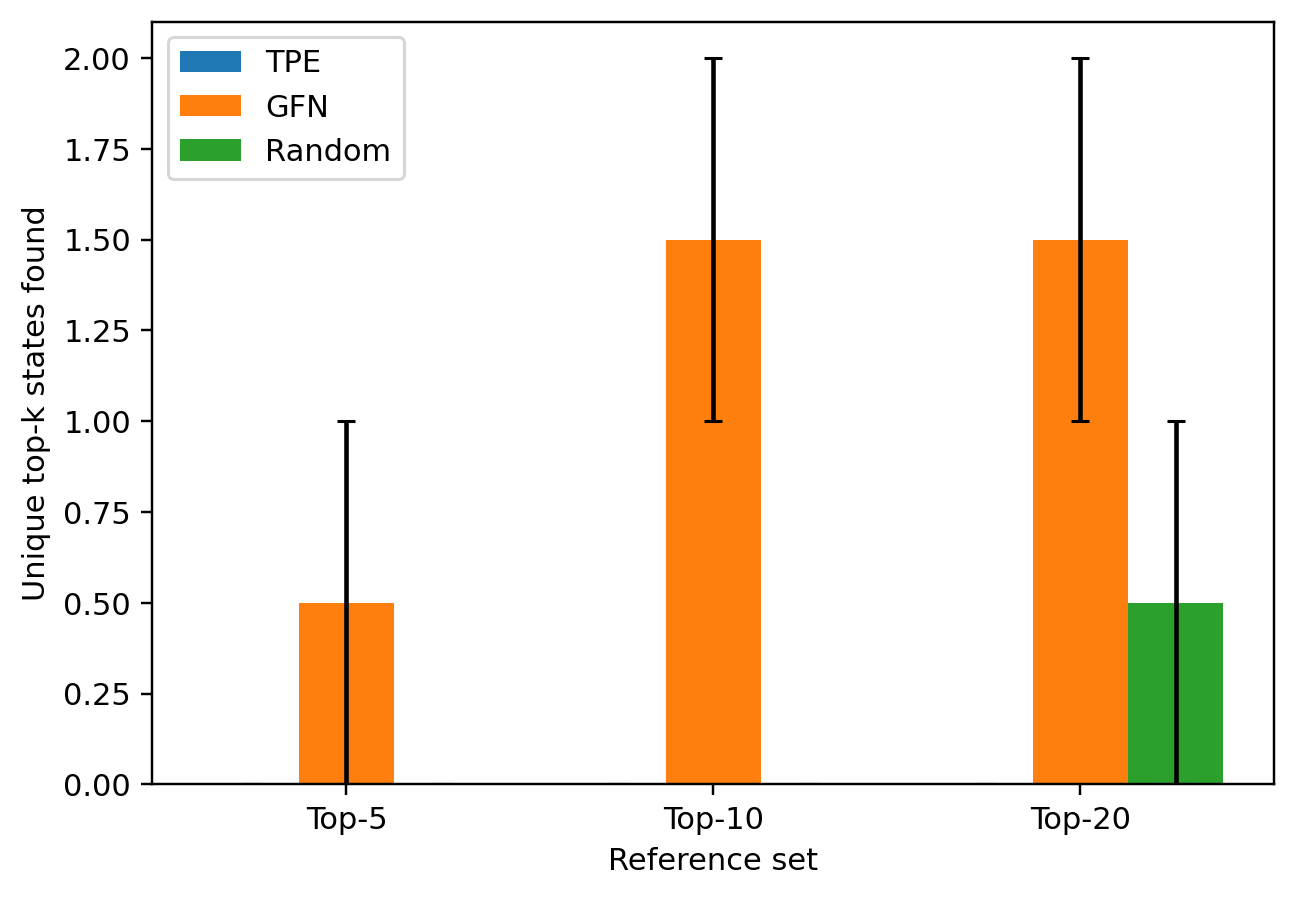}
        
        \small (b) \(sf=0.3,\ \beta=4\)
    \end{minipage}
    \hfill
    \begin{minipage}[t]{0.32\textwidth}
        \centering
        \includegraphics[width=\linewidth]{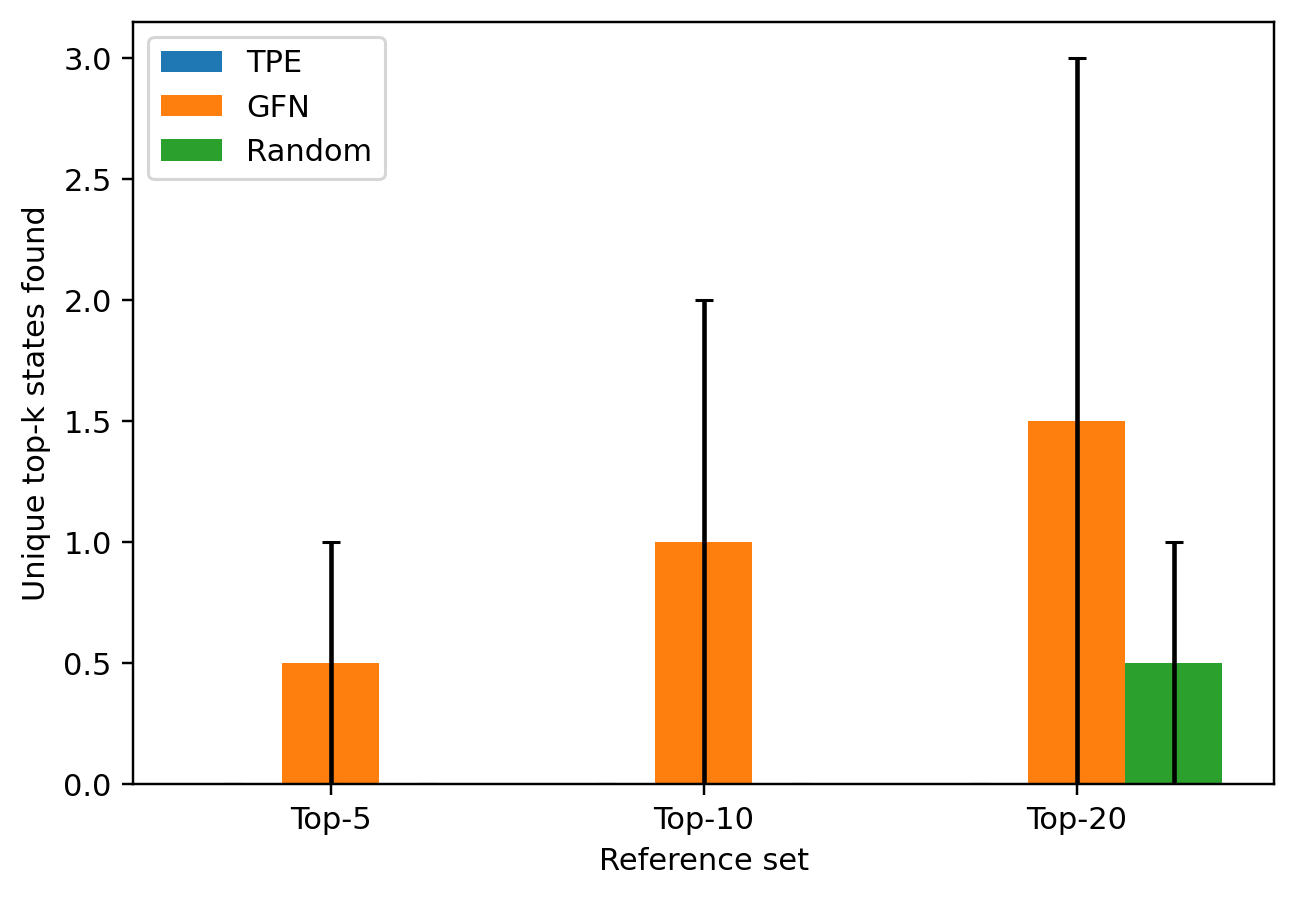}
        
        \small (c) \(sf=0.3,\ \beta=8\)
    \end{minipage}
    \caption{Recovery of unique target top-\(k\) states in the enumerable 1-cycle setting for the multimodal \(sf=0.3\) regime under three reward sharpness settings.}
    \label{fig:rq2_topk_recovery}
\end{figure}

This distributional agreement is reflected in retrieval behavior. \Fig{fig:rq2_topk_recovery} measures how many distinct target top-\(k\) states are recovered by each strategy in the representative multimodal \(sf=0.3\) setting. Across all three \(\beta\) values, the GFlowNet consistently recovers more unique target top-\(k\) states than the competing methods. By contrast, the tree-structured Parzen estimator (TPE) recovers little or none of the exact target top-\(k\) set in this setting, while random search recovers only a limited number of top states at larger \(k\). This is important because it shows that the learned distribution is not merely similar to the target in aggregate. It is also useful in practice for retrieving states that belong to the exact high-quality subset of the landscape.

\begin{figure}[H]
    \centering
    \includegraphics[width=\linewidth]{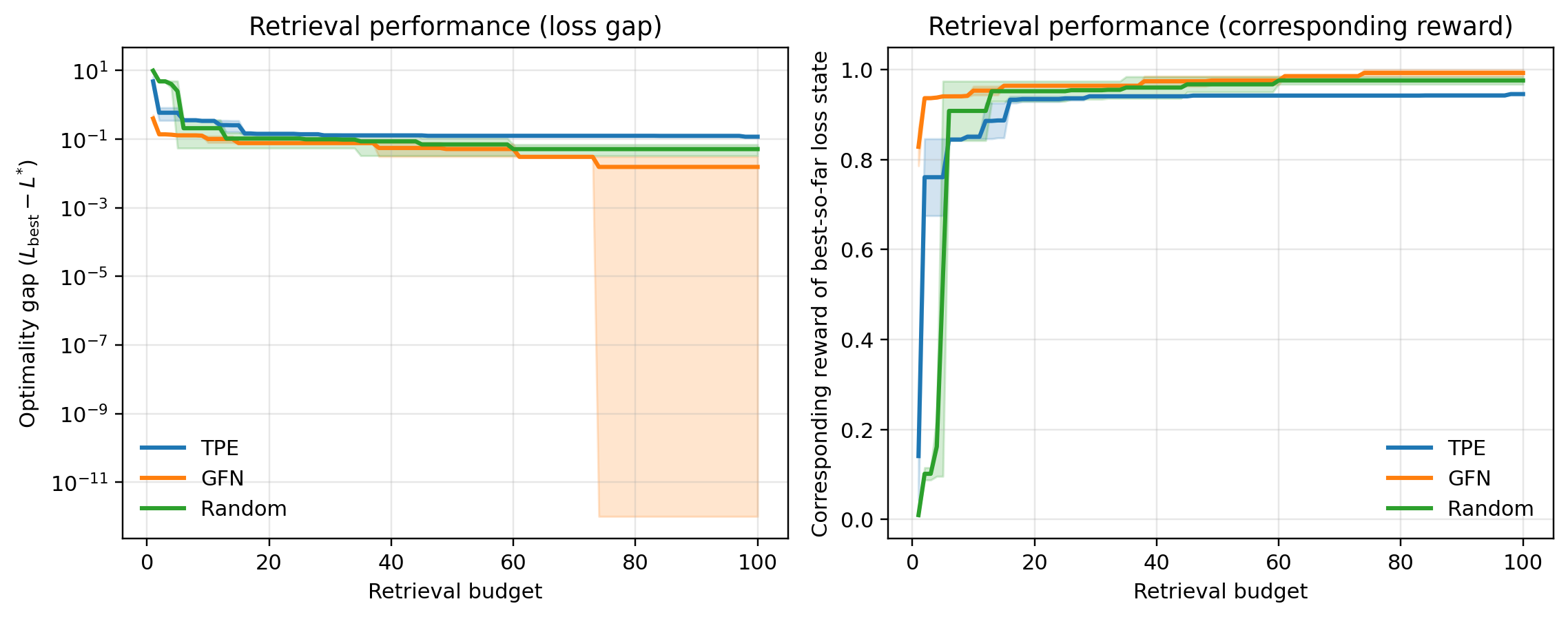}
    \caption{Retrieval performance in the enumerable 1-cycle setting for \(sf=0.3\) and \(\beta=4\). The left panel reports the best-so-far optimality gap, while the right panel reports the corresponding reward of the best-so-far state. Curves are aggregated across seeds.}
    \label{fig:rq2_retrieval_curves}
\end{figure}

The retrieval curves in \Fig{fig:rq2_retrieval_curves} provide a budget-aware view of the same phenomenon in the representative multimodal setting \(sf=0.3,\beta=4\). In this setting, the GFlowNet reaches strong configurations rapidly and remains competitive throughout the evaluation budget. At the same time, random search also performs surprisingly well. This should be interpreted jointly with \Fig{fig:rq2_probability_rank_compare}. The target distribution in this setting remains relatively flat, so many terminal states have similar losses. Combined with the size of the enumerable state space, this makes strong configurations comparatively easy to encounter by uninformed sampling alone, which in turn favors random search in the retrieval curves. The idea is not that random search reveals useful structure in the landscape, but that the learned policy remains competitive even in a regime where the landscape offers a comparatively weak separation between strong and mediocre candidates.

\begin{figure}[H]
    \centering

    \begin{minipage}[t]{0.99\textwidth}
        \centering
        \includegraphics[width=\linewidth]{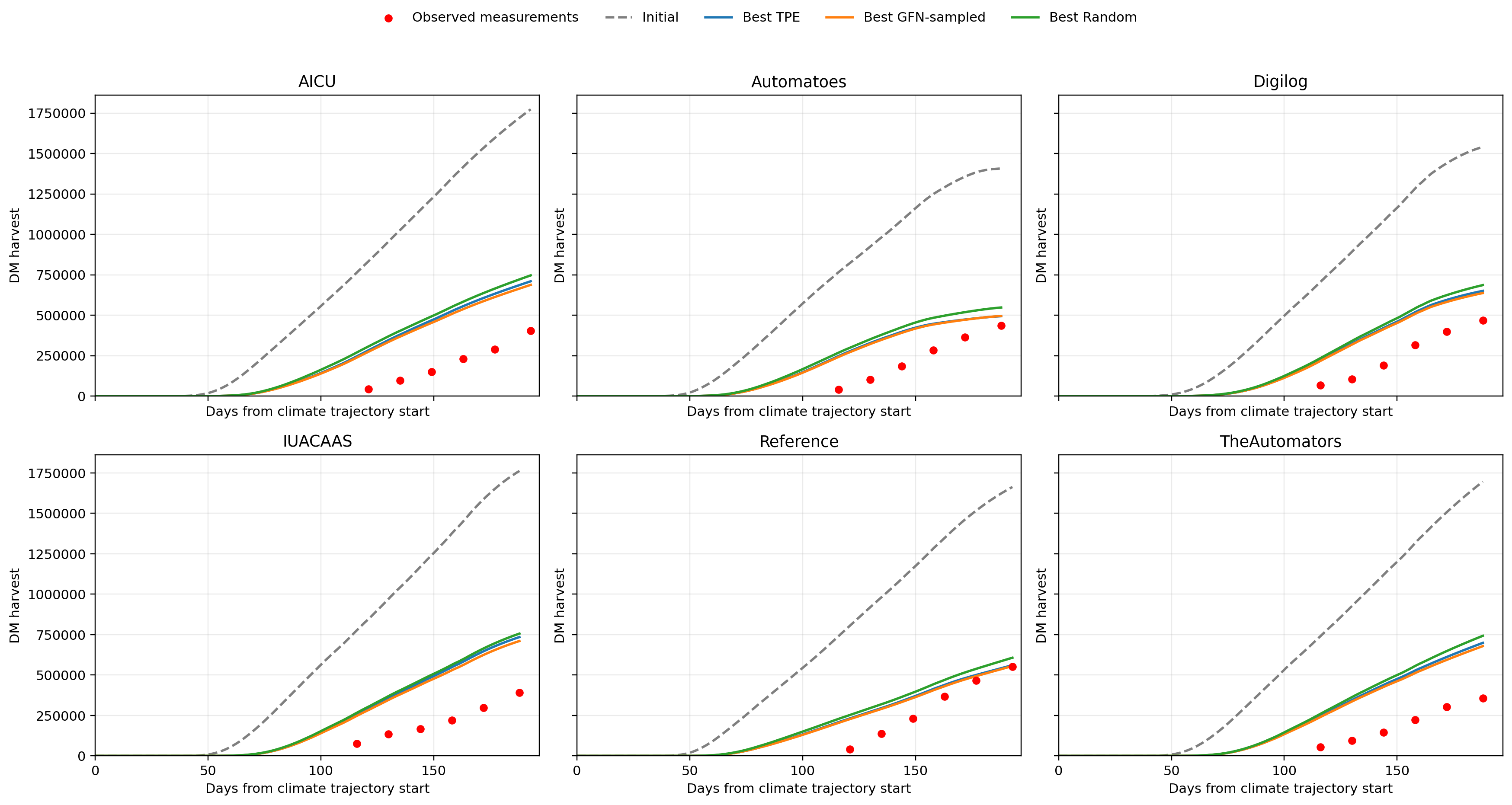}
        
        \small (a) Best-state overlay for \(sf=0.15,\ \beta=4\)
    \end{minipage}
    \hfill
    \begin{minipage}[t]{0.99\textwidth}
        \centering
        \includegraphics[width=\linewidth]{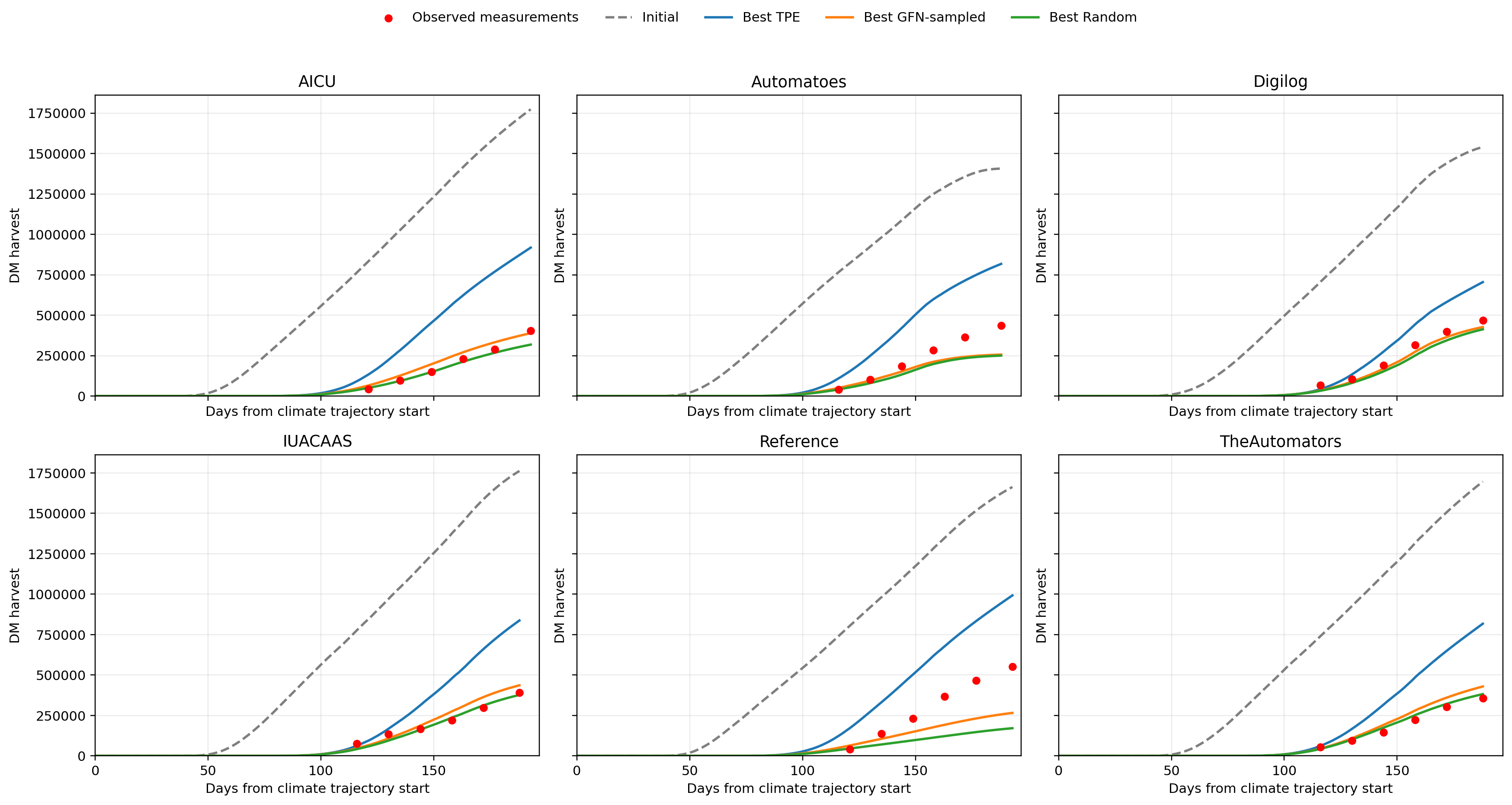}
        
        \small (b) Best-state overlay for \(sf=0.3,\ \beta=4\)
    \end{minipage}

    \caption{Best-state overlays in the enumerable 1-cycle setting for the representative \(\beta=4\) case under two step fractions. For each team, observed dry-mass measurements are compared against the trajectories produced by the best retrieved state of each method and by the initial configuration.}
    \label{fig:rq2_best_state_overlays}
\end{figure}

Finally, \Fig{fig:rq2_best_state_overlays} reconnects the retrieval results to the controlled environment agriculture setting. Rather than evaluating states only through ranked probabilities or loss values, these overlays show how the best configurations retrieved by each method reproduce the observed dry-mass measurements across teams. In both settings, the initial configuration remains clearly misaligned with the observations, while the retrieved states move the simulator trajectories toward the measured behavior. At \(sf=0.15\), the differences between retrieved trajectories remain relatively modest, whereas the \(sf=0.3\) setting makes the contrast clearer: the retrieved configurations remain much closer to the observations than the initial state, while the adaptive baseline tends to overshoot cumulative dry mass more strongly. In that sense, the figure provides an interpretable bridge between state-space retrieval quality and the physical-system trajectories that motivate adaptation.

Taken together, these results show that the learned policy is not only capable of reproducing the dominant structure of the target distribution, but is also useful as a retrieval mechanism for strong simulator parameterizations. The close agreement in ranked probability profiles, the repeated recovery of target top-\(k\) states, and the improved best-so-far performance all indicate that the GFlowNet supports \rone{} in practice: it remains sufficiently grounded in the adaptation landscape to retrieve parameterizations that are well aligned with the observations. At the same time, because this retrieval performance is achieved through a learned distribution rather than through a single deterministic calibration, the approach satisfies \rtwo{}.

\subsubsection{Scalability}
\label{sec:results-rq3}
\textbf{RQ3} explores how the proposed approach behaves when the adaptation problem is scaled beyond the enumerable 1-cycle regime. To study this question, we consider the 2-cycle setting, where the number of reachable terminal configurations increases substantially and the exact target distribution is no longer available. In this regime, the evaluation focuses on retrieval quality, diversity among the strongest retrieved candidates, and qualitative agreement with the observed trajectories rather than on exact distributional fidelity. We restrict the analysis to \(\beta=4\). This choice was motivated by the need to keep the reward landscape smooth enough to remain conducive to learning while still sharp enough to retrieve high-reward samples.

\begin{figure}[H]
    \centering

    \begin{minipage}[t]{0.78\textwidth}
        \centering
        \includegraphics[width=\linewidth]{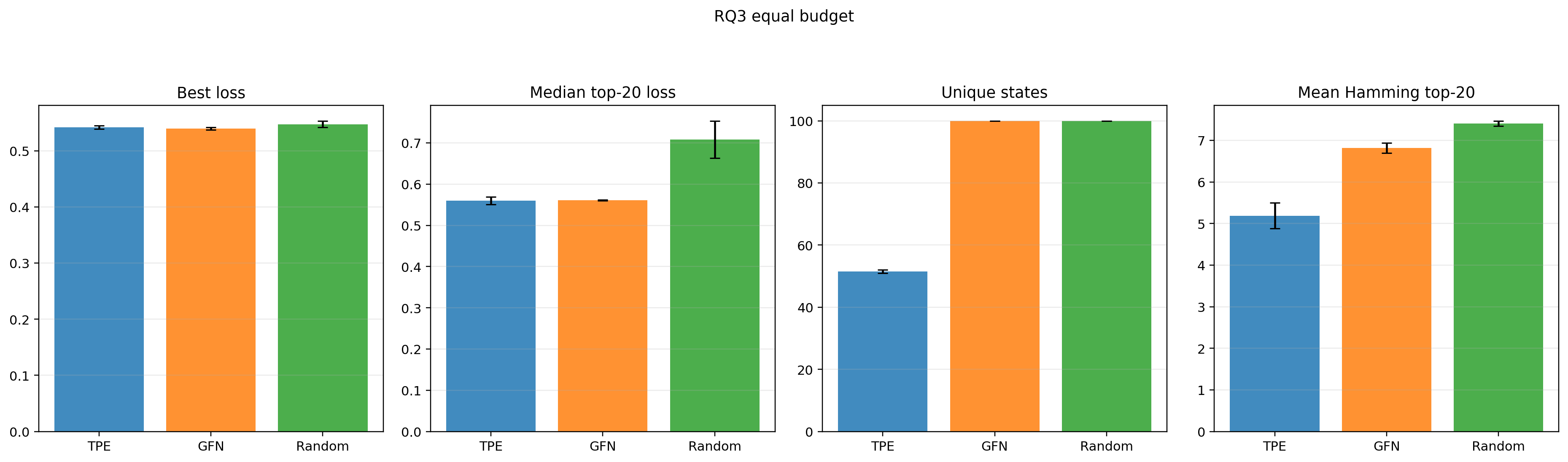}
        
        \small (a) Equal-budget comparison for \(sf=0.15,\ \beta=4\)
    \end{minipage}

    \vspace{0.8em}

    \begin{minipage}[t]{0.78\textwidth}
        \centering
        \includegraphics[width=\linewidth]{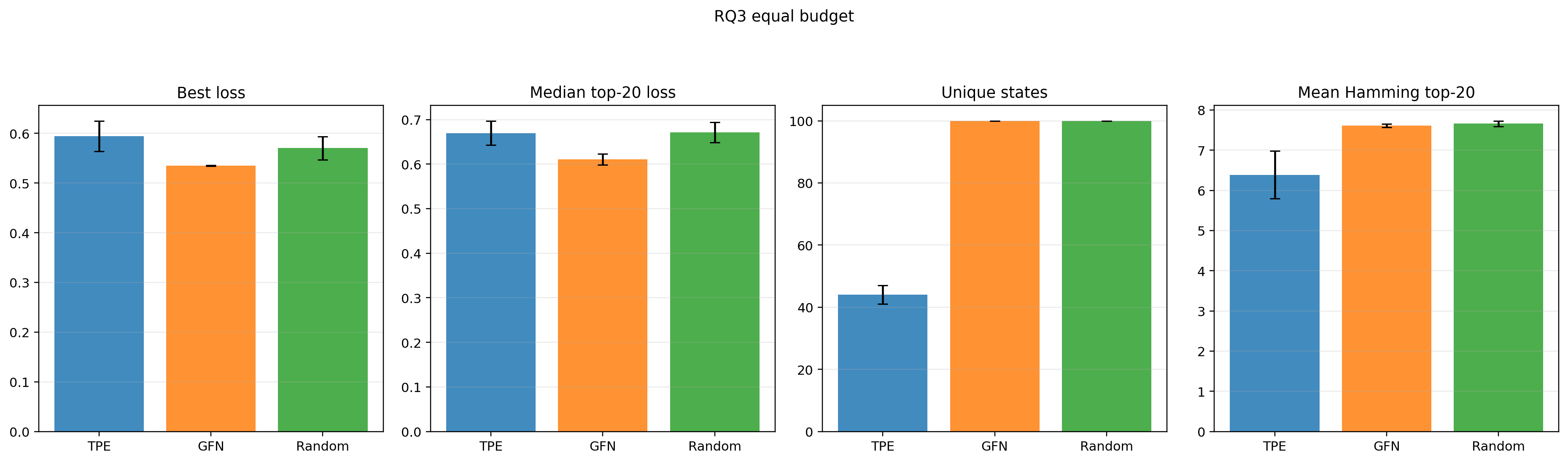}
        
        \small (b) Equal-budget comparison for \(sf=0.3,\ \beta=4\)
    \end{minipage}

    \caption{Equal-budget comparison in the 2-cycle setting for the representative \(\beta=4\) case under two step fractions. Best loss and median top-20 loss evaluate the quality of the retrieved states, while the mean Hamming distance among the top retrieved states measures how broadly each method explores distinct high-quality regions of the search space.}
    \label{fig:rq3_equal_budget_compare}
\end{figure}

\Fig{fig:rq3_equal_budget_compare} evaluates retrieval quality and diversity after the same simulation budget. The two loss-based quantities capture complementary aspects of retrieval quality. Best loss measures the quality of the single strongest state found by each method, whereas median top-20 loss indicates whether good performance is limited to one isolated configuration or extends to a broader set of strong candidates. The mean Hamming distance among the top retrieved states measures diversity: larger values indicate that the strongest retrieved candidates are distributed across more distinct regions of the search space rather than concentrated around near-duplicate solutions, since their keys differ by a greater amount.

Under this equal-budget comparison, the GFlowNet remains competitive with TPE on the quality measures while maintaining a substantially broader spread of high-quality candidates. At \(sf=0.15\), the two methods remain close in terms of best loss and median top-20 loss, but the Hamming distance clearly separates them, showing that the learned policy explores a more diverse set of promising configurations. At \(sf=0.3\), the GFlowNet becomes more favorable on the quality measures as well, while still preserving this broader diversity. This indicates that, even in the larger 2-cycle regime, the learned policy does not collapse onto a narrow subset of configurations, but continues to retrieve strong candidates from multiple regions of the parameter space.

At the same time, random search performs strongly in both settings, and especially in the coarser \(sf=0.3\) regime. This should not be read as evidence that uninformed sampling is intrinsically competitive with a learned retrieval policy, but rather as a consequence of the enlarged 2-cycle state space and of the limited simulator budget available to train the GFlowNet. In this regime, the learned policy can still retrieve states with lower loss than random search, but it no longer separates itself as clearly as in the enumerable 1-cycle setting because it has not amortized enough of the larger search space. This is reflected in the diversity results as well. Random search attains the largest Hamming distances, which is expected from an uninformed sampling strategy, while the GFlowNet achieves comparable diversity with consistently better loss values. Read in that light, the main result is not that random search reveals useful structure in the landscape, but that the learned policy remains competitive even though the enlarged search space makes it much harder to improve decisively over a flat reward landscape.

\begin{table}[htbp]
    \centering
    \resizebox{\textwidth}{!}{%
    \begin{tabular}{|c|c|c|c|c|c|}
        \hline
        Step fraction & Method & Best loss & Median top-20 loss & Mean Hamming top-20 & Training time \\
        \hline
        $sf=0.15$ & TPE
        & $0.542 \pm 0.003$
        & \textbf{\boldmath $0.560 \pm 0.009$}
        & $5.19 \pm 0.31$
        & \textbf{\boldmath$1{:}37{:}46 \pm 17{:}22$} \\
        $sf=0.15$ & GFN
        & \textbf{\boldmath $0.539 \pm 0.002$}
        & $0.561 \pm 0.001$
        & $6.81 \pm 0.12$
        & $12{:}51{:}33 \pm 2{:}14{:}28$ \\
        $sf=0.15$ & Random
        & $0.547 \pm 0.006$
        & $0.709 \pm 0.045$
        & \textbf{\boldmath $7.41 \pm 0.06$}
        & --- \\
        \hline
        $sf=0.3$ & TPE
        & $0.595 \pm 0.031$
        & $0.670 \pm 0.027$
        & $6.39 \pm 0.59$
        & \textbf{\boldmath$1{:}46{:}29 \pm 18{:}13$} \\
        $sf=0.3$ & GFN
        & \textbf{\boldmath $0.535 \pm 0.001$}
        & \textbf{\boldmath $0.611 \pm 0.012$}
        & $7.61 \pm 0.04$
        & $8{:}42{:}14 \pm 3{:}43{:}30$ \\
        $sf=0.3$ & Random
        & $0.570 \pm 0.024$
        & $0.671 \pm 0.022$
        & \textbf{\boldmath $7.66 \pm 0.07$}
        & --- \\
        \hline
    \end{tabular}
    }
    \caption{RQ3 summary in the 2-cycle setting under an equal retrieval budget for $\beta=4$. Lower is better for best loss and median top-20 loss; higher is better for mean Hamming top-20.}
    \label{tab:rq3_summary}
\end{table}

Table~\ref{tab:rq3_summary} highlights the main trade-off that emerges in the 2-cycle setting. At \(sf=0.15\), the GFlowNet only marginally improves upon TPE on best loss, while TPE attains a nearly identical median top-20 loss and fits about six times faster. At \(sf=0.3\), the distinction becomes clearer: the GFlowNet achieves the best values for both best loss and median top-20 loss, which indicates that it retrieves not only a strong single candidate, but also a stronger set of high-quality candidates under the same budget. The diversity column complements this comparison. Random search attains the largest Hamming distances in both settings, which is expected from an uninformed strategy that spreads samples broadly across the state space. More importantly, the GFlowNet maintains higher Hamming distances than TPE while remaining competitive, or favorable, on the quality measures. This suggests that the learned policy continues to cover several distinct promising regions of the search space rather than concentrating exclusively on one narrow optimum. At the same time, this broader coverage comes at a substantial computational cost: fitting the 2-cycle GFlowNet requires roughly 9 to 13 hours, whereas TPE typically completes in under 2 hours. During training, our GFlowNet caches candidates and their score, causing subsequent seeds to train faster, which explains the observed variance in training times.

\begin{figure}[H]
    \centering
    \begin{minipage}[t]{0.99\textwidth}
        \centering
        \includegraphics[width=\linewidth]{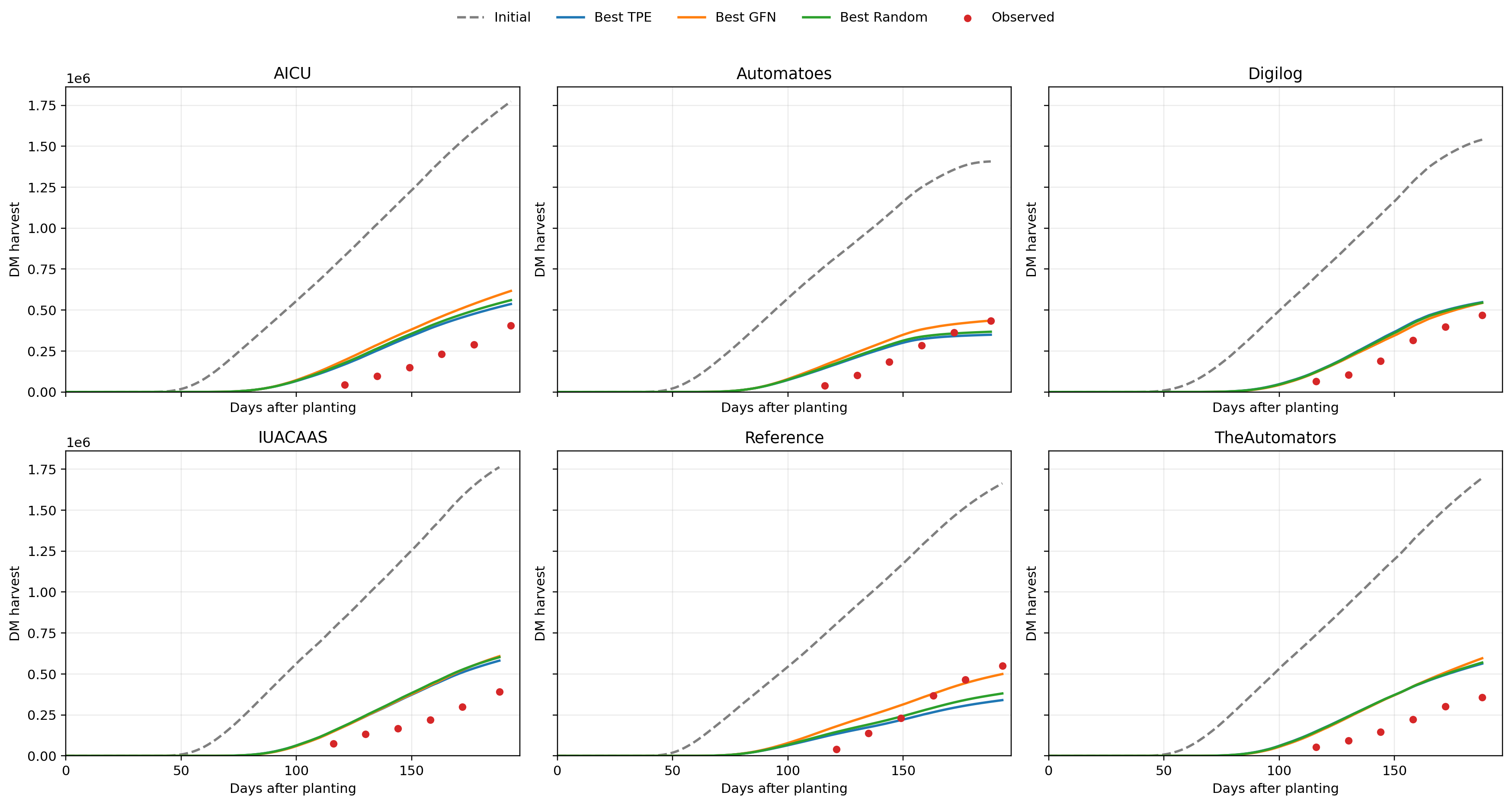}
        
        \small (a) Best-state overlay for \(sf=0.15,\ \beta=4\)
    \end{minipage}
    \hfill
    \begin{minipage}[t]{0.99\textwidth}
        \centering
        \includegraphics[width=\linewidth]{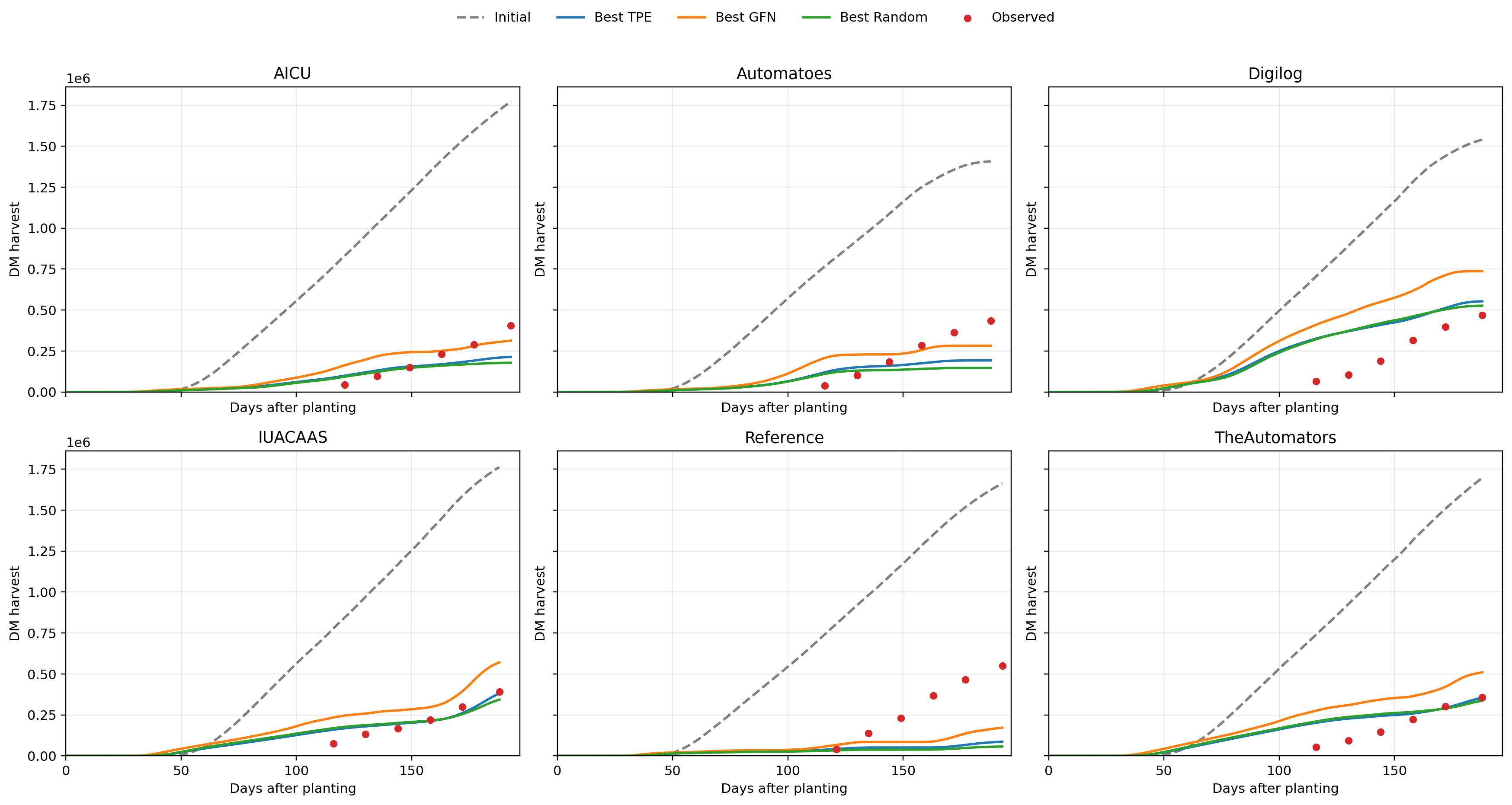}
        
        \small (b) Best-state overlay for \(sf=0.3,\ \beta=4\)
    \end{minipage}
    \caption{Best-state overlays in the 2-cycle setting for \(\beta=4\) under two step fractions. For each team, observed dry-mass measurements are compared against the trajectories produced by the best retrieved state of each method and by the initial configuration.}
    \label{fig:rq3_best_state_overlays}
\end{figure}

Finally, \Fig{fig:rq3_best_state_overlays} reconnects the scaling results to the greenhouse domain. Compared with the enumerable 1-cycle regime, the retrieved trajectories now display a much wider range of shapes across teams, which reflects the additional degrees of freedom introduced by the 2-cycle setting. In principle, this larger space should make better fits possible, since the simulator can express a broader set of cumulative dry-mass trajectories. In practice, however, this also makes the adaptation problem substantially harder to solve. In both settings, the retrieved configurations improve clearly over the initial state, which remains visibly misaligned with the observations, but the differences between methods become more modest and more team-dependent than in the 1-cycle case. At \(sf=0.15\), many of the retrieved trajectories remain fairly close to one another, which is consistent with the small separation observed in the equal-budget comparison. At \(sf=0.3\), the contrast becomes clearer, but no single method dominates uniformly across all teams. This supports the broader interpretation of \textbf{RQ3}: scaling increases the expressive flexibility of the search space, but with simulation as the main bottleneck, identifying those better-fitting regions reliably becomes a much more difficult problem.

Taken together, these results show that the proposed approach continues to support both \rone{} and \rtwo{} when the adaptation problem is scaled to the 2-cycle regime, but that its practical effectiveness is currently limited by amortization cost. Under a matched retrieval budget, the GFlowNet remains competitive and continues to preserve a broad set of plausible candidate states, which is consistent with \rtwo{}. At the same time, it remains capable of retrieving parameterizations that are well aligned with the observations, which supports \rone{}, albeit in a lesser capacity than the 1-cycle case. However, the time required to fit the learned policy remains substantial, and this now constitutes the main performance bottleneck of the approach. In that sense, the 2-cycle results do not contradict the generative objective of the method. They show instead that further improvements in evaluation efficiency, batching, or surrogate support will be needed before amortized generative retrieval can be fully exploited at this scale.

%% file: sections/experiments/discussion.tex
\subsection{Discussion}
\label{sec:discussion}
This study examined whether GFlowNet-based adaptation can satisfy \rone{} and \rtwo{} for digital twins of natural systems, where sparse and noisy observations often make simulator calibration underdetermined. The results suggest that the answer is positive in the enumerable 1-cycle regime, and only partially so at larger scale. In the 1-cycle setting, the learned policy recovered the dominant structure of the adaptation landscape, remained closely aligned with the exact target distribution, and retrieved parameterizations whose simulated trajectories matched the observations. In the 2-cycle setting, the same qualitative behavior persisted, but amortization costs grew substantially with the size of the search space.

The clearest evidence for \rtwo{} comes from \textbf{RQ1}. In the enumerable regime, the learned policy did not collapse onto a single optimum and instead preserved several plausible high-quality regions of the landscape. This property is essential for digital twins of natural systems, where multiple simulator parameterizations may remain consistent with the available evidence. Preserving several plausible candidates is therefore not incidental to adaptation, but part of what makes it useful for downstream tasks such as monitoring, prediction, and prescription.

Conversely, RQ2 provides the clearest support for \rone{}. In the 1-cycle setting, the ranked probability comparisons, top-\(k\) recovery results, retrieval curves, and best-state overlays all indicate that the learned policy remains sufficiently grounded in the adaptation landscape to recover parameterizations whose simulated trajectories are well aligned with the observations. This complements the interpretation of RQ1 directly. If RQ1 shows that the learned policy preserves the relevant structure of the landscape, RQ2 shows that this structure is not merely descriptive, but can be exploited to retrieve strong, observation-aligned configurations in practice.

\textbf{RQ3} then shows how both conclusions change when we increase the possible state space. In the 2-cycle regime, the learned policy continues to retrieve strong candidates and to maintain a broader spread of promising configurations than TPE, which suggests that \rtwo{} is retained at least in part in the enlarged search space. Support for \rone{}, however, becomes more conditional. The learned policy remains competitive under a matched retrieval budget, but its separation from TPE and from random search is substantially reduced. We interpret this not as a breakdown of the formulation itself, but as evidence that the available simulator budget is insufficient to amortize the enlarged landscape to the point where the retrieval advantage remains decisive.

The main limitation revealed by these results is computational rather than conceptual. The 2-cycle regime substantially increases the number of reachable terminal configurations, and simulator evaluation becomes the dominant bottleneck. In the present setup, fitting the GFlowNet required many hours of training, yet this investment was still insufficient to fully amortize the larger landscape. A second limitation concerns the induced reward landscape itself. When the landscape is relatively flat, many states achieve similar losses, which weakens the separation between informed and uninformed exploration and can make random search appear more competitive than it would under a sharper regime. This flatness also affects the learned sampler. If such a landscape were matched exactly, very high-reward states could remain rarely sampled simply because of the sheer size of the state space. The GFlowNet mechanism addresses this to some extent through proportional sampling over the reward landscape. As observed in \textbf{RQ2}, different values of \(\beta\) affected the norm of the learned distribution. By adjusting this norm, probability mass can be concentrated more strongly on top-ranked states, which helps support the sampling of high-reward calibration hypotheses in downstream digital twin services.

Even with these limitations, the results remain encouraging for digital twins of natural systems. They show that a generative adaptation service can preserve multiple plausible parameterizations while still retrieving strong ones, which is precisely the combination required when observations are destructive, costly, or sparsely distributed in time. The main challenge exposed by the present study is therefore efficiency: progress at larger scale will depend on better batching, stronger caching strategies, or surrogate support that enables more effective amortization of larger search spaces.

%% file: sections/experiments/threats.tex
\subsection{Threats to validity}
\label{sec:threat}
\textbf{Internal validity.} The reported results depend on several experimental choices. The adaptation landscape is induced by a specific grouped perturbation design, reward formulation, and simulator-based loss definition, all of which affect the relative difficulty of the retrieval problem for GFlowNet, TPE, and random search. In addition, the 1-cycle and 2-cycle regimes support different forms of evidence: the former benefits from an exact enumerable target distribution, whereas the latter can only be assessed through retrieval-based comparisons. The number of seeds and computational budgets also remains limited, and caching affects wall-clock time by reducing simulator effort unevenly across runs. Further, our experiments were performed on clusters with shared access across our institutions, thus computing power was not consistent throughout model training. Finally, flatter reward landscapes reduce the separation between strong and mediocre candidates, which can make random search appear more competitive than it would under a sharper target distribution.

\noindent\textbf{External validity.} The evaluation was conducted on a single mechanistic greenhouse simulator in controlled environment agriculture, with one grouped perturbation design, two step fractions, a narrow range of reward sharpness settings, and baselines centered on TPE and random search. The conclusions therefore support the feasibility of GFlowNet-based adaptation for digital twins of natural systems in this setting, but they do not establish that the same results will transfer unchanged to other simulators, crops, sensing regimes, or adaptation services. Broader generality will need to be assessed across other natural-system domains, other simulator structures, and larger search spaces where the computational bottlenecks identified here may manifest differently.

%% file: sections/background/related.tex
\section{Related Works}
\label{sec:related}
This section situates the proposed approach with respect to three complementary lines of work. We first review model-driven engineering research on digital twins, focusing on lifecycle management, adaptation, and the role of model artifacts and services. We then discuss digital twins of natural and biological systems, where uncertainty, indirect observations, and evolving operating contexts make simulator adaptation particularly challenging. Finally, we examine simulation-based inference and recent generative approaches for likelihood-free parameter inference in mechanistic models, in order to clarify how the present work differs from existing posterior-estimation and calibration-oriented formulations.

\subsection{Model-driven engineering for digital twins}

Systematic engineering of digital twins has emerged as a distinct MDE research agenda, with early work outlining its foundational challenges around model heterogeneity, synchronization across abstractions, and lifecycle-wide coordination~\cite{bordeleau2020towards}. Recent synthesis work extends this agenda by aligning MDE capabilities with digital twin engineering needs across the lifecycle~\cite{michael2025mde}, by providing a systematic mapping of MDE techniques applied to digital twins~\cite{lehner2025mde}, and by mapping the broader software engineering landscape~\cite{dalibor2022crossdomain}. On lifecycle management and self-adaptation, declarative formulations treat the twin and its physical counterpart as a coupled system evolving across lifecycle stages~\cite{kamburjan2024declarative}, while notational approaches reason about continuous digital twin evolution~\cite{mertens2024dartwin}. Our work contributes to this line of research by treating model adaptation as an explicit digital twin service and as the inference mechanism that refreshes the mechanistic simulator when new evidence justifies it, while locating this service within the design phase of the digital twin lifecycle.

\subsection{Digital twins of natural and biological systems}

Digital twins of natural systems remain less established but are directly relevant to our setting. Prior work in greenhouse-based cyber-biophysical systems identifies simulator construction, uncertainty handling, and adaptation as central challenges~\cite{david2023digital}, and has produced self-adaptation architectures that distinguish behavioral from structural adaptation~\cite{kamburjan2024greenhousedt}. Our contribution sits in the behavioral layer and replaces pointwise parameter tuning with reward-proportional sampling over complete simulator configurations. In smart farming more broadly, control-oriented digital twin frameworks have been proposed~\cite{verdouw2021smartfarming}, though recent reviews report that robust predictive capabilities remain limited in practice~\cite{purcell2023agriculture}. In biomedicine, digital twins have been grounded in probabilistic graphical models~\cite{kapteyn2021probabilistic}, with uncertainty quantification discussed as a central concern across the lifecycle~\cite{thelen2023comprehensive}. Despite their domain differences, these systems share a common modeling difficulty: observations are typically partial, heterogeneous, and obtained at the level of system behavior rather than directly at the level of simulator parameters. Maintaining several plausible simulator configurations is thus valuable because it preserves adaptation uncertainty in a form that can still support downstream prediction, what-if analysis, and decision-making.

\subsection{Simulation-based inference and generative modeling}

Calibration under intractable likelihoods is typically addressed through approximate Bayesian computation~\cite{sisson2018handbook} or neural posterior estimation~\cite{greenberg2019npe}, as surveyed in~\cite{cranmer2020frontier}. Recent work applies these techniques directly to digital twins, using neural posterior estimation for amortized parameter inference in physiological digital twins~\cite{hoang2025real} and formulating probabilistic digital twins through measure-theoretic constructions~\cite{agrell2023optimal}. These approaches target posterior estimation over parameters. We instead learn a generative policy that constructs complete simulator configurations, preserving the compositional structure of the mechanistic model. Generative Flow Networks provide the formal basis for reward-proportional sampling of compositional objects, with mode-covering behavior that distinguishes them from reward-maximizing reinforcement learning and mode-seeking variational inference, and have been surveyed as tools for scientific discovery~\cite{jain2023gflownets}. Established applications include Bayesian structure learning~\cite{deleu2022bayesian} and multi-objective optimization~\cite{jain2023multiobj}, with training stabilized through trajectory balance~\cite{malkin2022trajectory}. To our knowledge, prior work has not applied GFlowNets to mechanistic simulator adaptation within a digital twin setting.

%% file: sections/conclusion/conclusion.tex
\section{Conclusion}
\label{sec:conclusion}

This work investigates model adaptation as a service within the lifecycle of digital twins of natural systems, with generative modeling as an inference mechanism. The objective was not simply to recalibrate a mechanistic simulator, but to maintain its operational usefulness as new evidence becomes available. To that end, the adaptation service was framed around two requirements: \rone{}, the ability to retrieve parameterizations that remain well aligned with the available observations, and \rtwo{}, the ability to preserve several plausible high-quality parameterizations rather than collapse prematurely onto a single solution. These requirements are especially important in natural systems, where observations are sparse, noisy, and often insufficient to identify one unambiguous explanation of the physical process.

Our results show that this objective is attainable when the parameterization is characterized by a small state space. The learned model policy recovered the dominant structure of the adaptation landscape, remained closely aligned with the exact target distribution, and retrieved parameterizations whose simulated trajectories matched the observations. In that setting, the method provided strong support for both \rone{} and \rtwo{}: it remained grounded enough to recover observation-aligned configurations while also preserving several plausible alternatives that can support downstream digital twin services.

Results on a larger parameter state-space showed that achieving better performance is limited by computational costs. Under a matched retrieval budget, the GFlowNet remained competitive and continued to preserve a broader spread of plausible candidates than baselines. However, the enlarged search space made amortization substantially more difficult, and simulator evaluation emerged as the dominant bottleneck. The main limitation identified by this study is thus not conceptual, but computational.

Taken together, our results suggest that GFlowNets are a promising foundation for model adaptation in digital twins of natural systems. They support both retrieval of observation-aligned parameterizations and preservation of plausible alternatives, which is precisely the combination required in underdetermined natural-system settings. Future work should therefore focus on improving scalability through more efficient simulator use.